\documentclass[sigconf]{acmart}
\usepackage{multirow}
\usepackage{makecell}
\usepackage{colortbl}
\usepackage{overpic}
\usepackage{amsmath}
\usepackage{empheq}
\usepackage{comment}
\usepackage{makecell}
\usepackage{multirow}
\usepackage{multicol}
\usepackage{float}
\usepackage{marvosym}
\usepackage{balance}
\definecolor{tabfirst}{rgb}{1, 0.7, 0.7} 
\definecolor{tabsecond}{rgb}{1, 0.85, 0.7} 
\definecolor{tabthird}{rgb}{1, 1, 0.7} 
\usepackage{float}

\AtBeginDocument{%
  }

\copyrightyear{2025}
\acmYear{2025}
\setcopyright{acmlicensed}
\acmConference[ICMR '25]{Proceedings of the 2025 International Conference on Multimedia Retrieval}{June 30-July 3, 2025}{Chicago, IL, USA}
\acmBooktitle{Proceedings of the 2025 International Conference on Multimedia Retrieval (ICMR '25), June 30-July 3, 2025, Chicago, IL, USA}
\acmDOI{10.1145/3731715.3733288}
\acmISBN{979-8-4007-1877-9/2025/06}

\begin{document}

\title{ClothHMR: 3D Mesh Recovery of Humans in Diverse Clothing from Single Image}

\author{Yunqi Gao}
\orcid{0009-0004-4194-665X}
\affiliation{%
  \institution{Central China Normal University}
  \city{Wuhan}
  \state{Hubei}
  \country{China}
}
\email{gaoyunqi@mails.ccnu.edu.cn}
\author{Leyuan Liu}
\orcid{0000-0002-8050-8677}
\authornote{Corresponding authors}
\affiliation{%
  \institution{Central China Normal University}
  \city{Wuhan}
  \state{Hubei}
  \country{China}
}
\email{lyliu@mail.ccnu.edu.cn}

\author{Yuhan Li}
\orcid{0009-0008-8256-6281}
\affiliation{%
  \institution{Huazhong University of Science and Technology}
  \city{Wuhan}
  \state{Hubei}
  \country{China}
}
\email{yuhanli@hust.edu.cn}
\author{Changxin Gao}
\orcid{0000-0003-2736-3920}
\affiliation{%
  \institution{Huazhong University of Science and Technology}
  \city{Wuhan}
  \state{Hubei}
  \country{China}
}
\email{cgao@hust.edu.cn}

\author{Yuanyuan Liu}
\orcid{0000-0002-1913-0089}
\affiliation{%
  \institution{China University of Geosciences (WuHan)}
  \city{Wuhan}
  \state{Hubei}
  \country{China}
}
\email{liuyy@cug.edu.cn}
\author{Jingying Chen}
\orcid{0000-0002-1523-3478}
\affiliation{%
  \institution{Central China Normal University}
  \city{Wuhan}
  \state{Hubei}
  \country{China}
}
\email{chenjy@mail.ccnu.edu.cn}




\renewcommand{\shortauthors}{Yunqi Gao et al.}

\begin{abstract}
With 3D data rapidly emerging as an important form of multimedia information, 3D human mesh recovery technology has also advanced accordingly.
However, current methods mainly focus on handling humans wearing tight clothing and perform poorly when estimating body shapes and poses under diverse clothing, especially loose garments.
To this end, we make two key insights: (1) tailoring clothing to fit the human body can mitigate the adverse impact of clothing on 3D human mesh recovery, and (2) utilizing human visual information from large foundational models can enhance the generalization ability of the estimation. Based on these insights, we propose ClothHMR, to accurately recover 3D meshes of humans in diverse clothing. ClothHMR primarily consists of two modules: clothing tailoring (CT) and FHVM-based mesh recovering (MR). The CT module employs body semantic estimation and body edge prediction to tailor the clothing, ensuring it fits the body silhouette. The MR module optimizes the initial parameters of the 3D human mesh by continuously aligning the intermediate representations of the 3D mesh with those inferred from the foundational human visual model (FHVM).
ClothHMR can accurately recover 3D meshes of humans wearing diverse clothing, precisely estimating their body shapes and poses. Experimental results demonstrate that ClothHMR significantly outperforms existing state-of-the-art methods across benchmark datasets and in-the-wild images. Additionally, a web application for online fashion and shopping powered by ClothHMR is developed, illustrating that ClothHMR can effectively serve real-world usage scenarios.
The code and model for ClothHMR are available at: \url{https://github.com/starVisionTeam/ClothHMR}.

\end{abstract}


\begin{teaserfigure}
  \centering
\includegraphics[width=\textwidth]{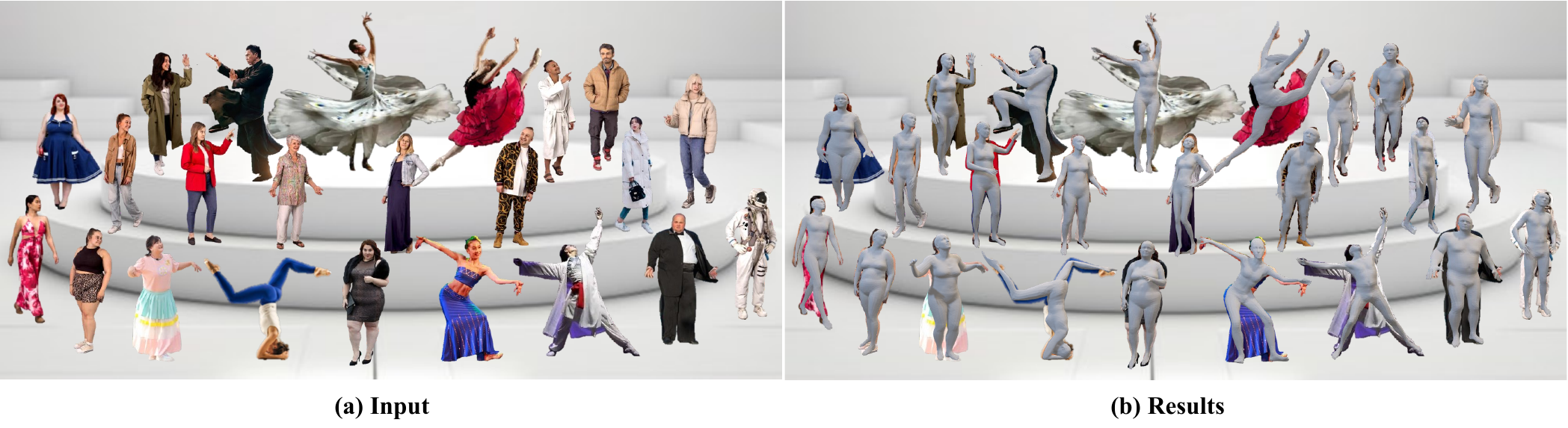}
  \caption{3D human meshes recovered by ClothHMR. ClothHMR can recover precise 3D meshes of humans wearing diverse clothing and in complex poses from a single image. Please zoom in to see the details.}
  \Description{}
  \label{Fig:show}
\end{teaserfigure}


\maketitle

\section{Introduction}

The widespread application of 3D data in multimedia has driven the rapid development of 3D human mesh recovery technology. This technology provides more realistic interactive experiences in fields such as virtual fitting, online fitness, and immersive gaming.
Although existing 3D human mesh recovery methods~\cite{survey} have made promising progress, they still face challenges when dealing with individuals wearing diverse garments, especially loose clothing.

\begin{figure}[h]
  \includegraphics[width=\linewidth]{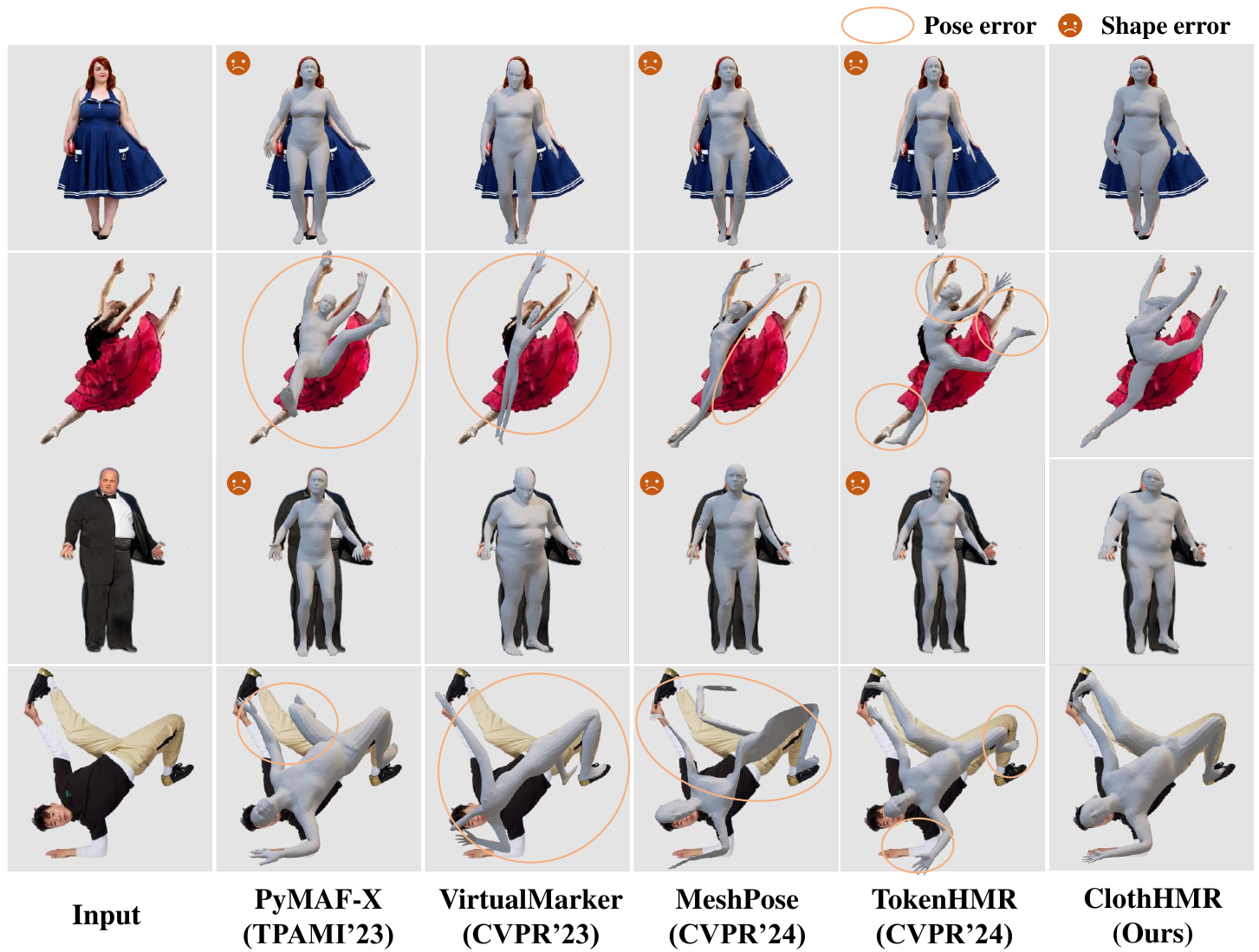}
  \caption{Summary of SOTA. 
  Current SOTA methods often misestimate body shapes or fail to handle complex poses when dealing with individuals wearing loose clothing. 
Even for individuals in non-loose clothing, current SOTA methods may still estimate poses without sufficient precision.
ClothHMR can effectively handle challenges posed by diverse types of clothing and complex poses.}
  \label{Fig:SOTA_summary}
\end{figure}

Due to the scarcity of annotated datasets with diverse clothing, few 3D human mesh recovery methods adequately account for the impact of clothing occlusion. 
As shown in Fig.~\ref{Fig:SOTA_summary}, current SOTA methods~\cite{pymaf,pymaf-X, Virtual_Markers, tokenhmr} often misestimate body shapes or fail to handle complex poses when dealing with individuals wearing loose clothing. 
Current SOTA methods may still estimate poses and shapes without sufficient precision even for individuals in non-loose clothing.

Instead of recovering 3D human meshes from raw RGB images directly, most methods utilize intermediate representations such as heatmaps~\cite{tung,neural,Pavlakos,STRAPS,Inter-part,ProHMR,Sim2real}, joints~\cite{HMR,Pose2Mesh,pymaf,pymaf-X,SMPLify,Zanfir}, silhouettes~\cite{Pavlakos,skeleton2mesh,Zanfir}, body segmentation~\cite{neural,NBF,chained}, and depth information~\cite{hdnet}.
Introducing intermediate representation distinctly contributes to overcoming data scarcity, thus guiding 3D human mesh recovery methods toward more accurate prediction~\cite{survey}.
However, inferring intermediate representations from raw RGB images remains an open problem that is not yet fully solved.
Any failure in intermediate representation inferring may lead to 3D human mesh recovery failures.
To address this, many methods~\cite{Virtual_Markers,zanfir2020weakly,3DCrowdNet,PaMIR,ICON,Unite_the_People,hdnet} combine multiple intermediate representations. However, the intermediate representations they employed are predicted by multiple deep neural networks, and the inconsistencies among these representations can compromise the recovery of 3D human meshes.
To improve the accuracy and robustness of 3D human mesh recovery, it is essential to utilize multiple intermediate representations that are both high-precision and consistent.

In this paper, we propose ClothHMR to recover 3D meshes of individuals wearing diverse clothing.
ClothHMR is based on two intuitive yet effective insights: (1) Tailoring clothing to fit the body can mitigate the adverse impact of clothing on 3D human mesh recovery;
(2) Intermediate representations extracted through a single foundational human vision model (FHVM) are more consistent and of higher quality.
To this end, ClothHMR primarily consists of two modules: clothing tailoring (CT) and FHVM-based mesh recovering (MR). 
Specifically, given a single RGB image, the CT module trims the clothing to fit the body, resulting in the tailored image. This module carries out body semantic estimation and body edge prediction to effectively trim excess clothing. The tailored image is then used to generate a 3D body mesh in the MR module. This module utilizes the foundational human visual model (FHVM) to extract three types of intermediate representations: joints, depth, and silhouette from the tailored image. It iteratively optimizes the 3D human mesh by minimizing the differences between the intermediate representations rendered from the estimated 3D human mesh and those extracted from the FHVM.

As shown in Fig.~\ref{Fig:show}, ClothHMR can recover precise 3D meshes of humans wearing diverse clothing and in complex poses from a single image.

Our contributions are summarized as follows:

(1) We propose a clothing-tailoring scheme, enabling 3D human mesh recovery algorithms to tackle individuals wearing diverse clothing, especially those dressed in loose clothing.

(2) We are the first to introduce a foundational human vision model into 3D human mesh recovery to enhance the precision of intermediate representations, thereby enabling the recovery of high-precision 3D human meshes using a simple iterative algorithm.

(3)  We propose ClothHMR, which can recover 3D meshes of individuals wearing diverse clothing, including extremely loose garments. ClothHMR significantly outperforms existing state-of-the-art methods by a substantial margin on both benchmark datasets and in-the-wild images.

\section{Relate Work}
\begin{figure*}[h]
  \includegraphics[width=\textwidth]{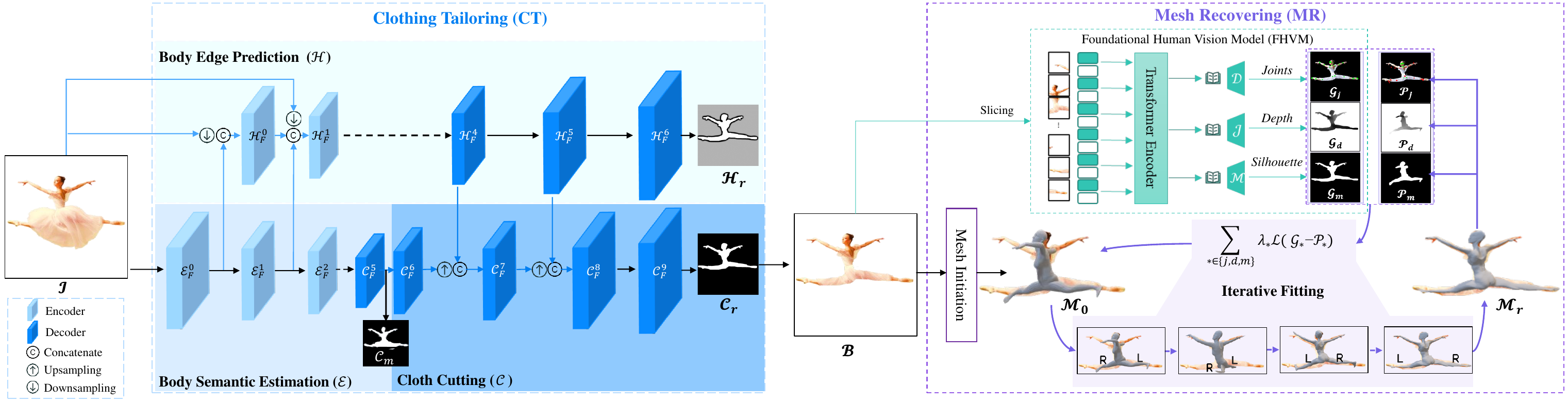}
  \caption{Overview. Given the image \(\mathcal{I}\), ClothHMR first uses the clothing tailoring module to trim the clothing to fit the human silhouette, resulting in \(\mathcal{B}\). Then, it initializes the mesh \(\mathcal{M}_0\) from \(\mathcal{B}\) and generates intermediate representations of \(\mathcal{B}\) using the foundational human vision model (FHVM), including joints \(\mathcal{G}_{j}\), depth \(\mathcal{G}_{d}\), and silhouette \(\mathcal{G}_{m}\). The mesh recovery module aligns the intermediate representations generated by FHVM with those produced by the mesh model (\(\mathcal{P}_{j}\), \(\mathcal{P}_{d}\), and \(\mathcal{P}_{m}\)), and through iterative optimization, ultimately generates an accurate human mesh \(\mathcal{M}_r\).  }
  \Description{}
  \label{Fig:framework}
\end{figure*}

\subsection{3D Human Mesh Recovery} 
3D human mesh recovery is represented using parametric human models, including the SCAPE model~\cite{SCAPE}, the SMPL model~\cite{smpl}, and its extension, SMPL-X~\cite{smpl-x}. The primary approaches can be divided into two categories: regression-based methods and optimization-based methods. 

Regression-based methods~\cite{HMR,DecoMR,Pose2Mesh,NBF,DSR,GCMR,ProHMR,HybrIK,DenseRaC,pymaf,3DCrowdNet,pymaf-X,Virtual_Markers,tokenhmr,meshpose,isair}, directly regress parameters from the networks and improve the accuracy of human pose and shape estimation through various forms of supervision. Kanazawa et al. propose HMR~\cite{HMR}, the first end-to-end regression method. Kolotouros et al. later introduced SPIN~\cite{SPIN}, which combines regression and optimization methods, significantly improving parameter estimation accuracy through an iterative optimization process. Zhang et al. further propose PyMAF~\cite{pymaf}, which explicitly corrects predicted parameters by recursively utilizing features within the deep regressor. They subsequently propose PyMAF-X~\cite{pymaf-X} to achieve more accurate hand representations.
Ma et al. propose VirtualMarker~\cite{Virtual_Markers} to estimate 64 joints on the human body surface for more accurate human pose estimation. Dwivedi et al. introduced tokenHMR~\cite{tokenhmr}, which addresses the inconsistency between 2D and 3D information caused by camera perspectives. 

Optimization-based methods~\cite{SMPLify,Unite_the_People,Zanfir,PaMIR,ICON,shapy,ShapeBoost,ms} focus on estimating 3D human meshes by minimizing the error between the mesh and 2D image information. Bogo et al. introduced SMPLify~\cite{SMPLify}, which iteratively fits the re-projection of SMPL model joints to those estimated from 2D images. Lassner et al.~\cite{Unite_the_People} enhanced this by adding human silhouettes for the re-projected mesh. Zanfir et al.~\cite{Zanfir} further improved the optimization by enforcing consistency between the re-projected mesh and semantic parts of the human body. Zheng et al.~\cite{PaMIR} propose PaMIR, which refines the estimation of parametric human model parameters through backward optimization using losses from 2D joints and clothed human reconstruction.

These methods generally achieve promising results when dealing with tight-fitting clothing. However, estimating human shape and pose under loose clothing remains a challenging problem.
Most existing human shape estimation methods rely on datasets such as Human3.6M~\cite{Human3.6M}, MPI-INF-3DHP~\cite{mpi-inf-3dhp}, 3DPW~\cite{3dpw}, COCO~\cite{coco}, and LSP~\cite{LSP}. Due to the lack of loose clothing data in these datasets, the accuracy of these methods decreases when handling loose clothing. ICON~\cite{ICON} attempts to mitigate the impact of clothing by ignoring the mask loss in regions with loose clothing, but it neglects the specific body shape beneath the loose clothing.

\subsection{Intermediate Representations}Intermediate representations, as simplified forms of RGB images, have been widely used in human mesh recovery. Common intermediate representations include 2D heatmaps~\cite{Virtual_Markers,3DCrowdNet,tung,neural,Pavlakos,STRAPS,zanfir2020weakly,Inter-part,ProHMR,Sim2real}, 2D/3D joints~\cite{HMR,Pose2Mesh,pymaf,pymaf-X,SMPLify,Unite_the_People,Zanfir,PaMIR,ICON}, silhouettes~\cite{Pavlakos,skeleton2mesh,Unite_the_People,Zanfir,PaMIR,ICON}, semantic segmentation~\cite{neural,NBF,zanfir2020weakly,chained}, and depth information~\cite{hdnet}.
VirtualMarkers~\cite{Virtual_Markers} retrieves human poses by multiplying joint heatmaps with the human mesh. HMR~\cite{HMR} uses 2D/3D joints to supervise the output of human mesh parameters. ICON~\cite{ICON} iteratively optimizes the human mesh using body silhouette information. NBF~\cite{NBF} takes semantic segmentation of the human body as input to predict the human model. HDNet~\cite{hdnet} utilizes depth information of the human body to reconstruct the clothed human.

However, traditional methods for estimating intermediate representations as~\cite{deepPose,openpose,alphapose,HigherHRNet,deeplabv3+,MODNet,hdnet} often suffer from a lack of generalization and high fidelity due to insufficient training data and simple network models. Therefore, there is an urgent need for foundational human vision models to enhance the quality of estimation of intermediate representations.

\subsection{Foundational Human Vision Models}
Computer vision has gradually adopted large-scale data for pre-training, enabling the learning of robust visual features~\cite{SLVM,MAE,VIT,AIM,dinov2}. The masked autoencoder (MAE)\cite{MAE} demonstrates excellent self-supervised learning capabilities by masking image patches and learning image representations during the reconstruction process. The vision transformer (ViT) enables efficient image processing by dividing images into fixed-size patches and applying transformer architecture to capture global relationships among these patches, enhancing performance in various vision tasks. 
With the advent of these pre-trained models, traditional research methods~\cite{deepPose,deeplabv3+,MODNet,hdnet} have also shifted, beginning to utilize pre-trained models for fine-tuning to accomplish specific tasks. MAWS~\cite{MAE1} studies MAE scaling on a billion images. ViTPose~\cite{xu2022vitpose} introduces a large-scale human pose estimation model based on ViT, significantly improving performance for human pose estimation tasks. Sapiens~\cite{sapiens} has introduced a series of high-resolution and large-scale ViT models, pre-trained on over 300 million in-the-wild human images. It supports 1K high-resolution inference and can easily adapt to various individual tasks, achieving high-fidelity inference in human vision tasks such as 2D pose estimation, human segmentation, depth estimation, and normal estimation.

\section{Method}


An overview of ClothHMR is illustrated in Fig.~\ref{Fig:framework}. Given a single RGB image as input, ClothHMR first employs the clothing tailoring (CT) module (Sec. ~\ref{sec:SE}) to tailor the clothing by trimming off the excess parts, then utilizes the mesh recovering (MR) module (Sec. ~\ref{sec:MR}) to generate the 3D body mesh that well-aligns with the cloth-tailored human image.
In the MR module, we first leverage a foundational human visual model (FHVM) to extract three kinds of intermediate representations, i.e., joints, depth, and silhouette, from the tailored image. We then iteratively optimize the 3D human mesh by minimizing the differences between the intermediate representations rendered from the estimated 3D human mesh and those extracted from the FHVM.

\subsection{Clothing Tailoring}
\label{sec:SE}
Clothing tailoring aims to precisely trim the garment to achieve a perfect fit with the human body shape. We designed a network consisting of three parts: body semantic estimation ($\mathcal{E}$), body edge prediction ($\mathcal{H}$), and cloth cutting ($\mathcal{C}$), which guides cloth cutting through the results of semantic estimation and edge prediction.

\noindent\textbf{Body Semantic Estimation.} The body semantic estimation first extracts deep features $\mathcal{E}_F^k(k={0,...,4})$ from the input image. The features $\mathcal{E}_F^k(k={0,...,2})$ will be effectively integrated into the edge prediction. To estimate the body semantic mask $\mathcal{C}_m$, we pass $\mathcal{E}_F^5$ through a convolutional layer followed by a Sigmoid activation, reducing its channel dimension to 1. The mask $\mathcal{C}_m$ is then supervised using a downsampled version of the ground truth matte \( \alpha_{g} \).
Since $\mathcal{C}_m$ is supposed to be smooth, we use the L2 loss as:
\begin{equation}
  {{L}_{cm}} = \frac{1}{2}{\left\| {{{\mathcal C}_m} - {G_{{\rm{16x}}}}({\alpha _g})} \right\|_2}
\end{equation}
where \( G_{16{\rm{x}}} \) denotes the operation involving \( 16 \times \) downsampling and Gaussian blurring \( \alpha_{g} \) represents the ground truth body semantics.

\noindent\textbf{Body Edge Prediction.} Body edge prediction aims to accurately mark the silhouette of the human body, which is critical for the subsequent steps in clothing cutting. To achieve precise identification, it integrates the input image, body semantic encoded features, and edge features as input. To fuse features from different resolutions, the image is downsampled using bilinear interpolation to produce images at various scales, denoted as ${{\mathcal I}_{1/s}}$.
\begin{equation}
\small
    {{\mathcal I}_{1/s}}(x,y) = \sum\limits_{i = 0}^{m - 1} {\sum\limits_{j = 0}^{n - 1} {\mathcal I} } (i,j) \cdot \frac{{(x - X(i))(y - Y(j))}}{{(X(i + 1) - X(i))(Y(j + 1) - Y(j))}}
\end{equation}
where ${\mathcal I}_{1/s}(x,y)$ represents the pixel value of the interpolated image at position \( (x, y) \). \( X(i) \) and \( Y(j) \) represent the coordinates corresponding to the position \( (i, j) \) in the original image, while \( X(i+1) \) and \( Y(j+1) \) are the coordinates of adjacent positions in the original image. When downsampling to $1/s$, the relationship between the new coordinates \( (x, y) \) and the original coordinates \( (i, j) \) is given by: $x = s \cdot X(i)$ and $y = s \cdot Y(j)$.
Then the feature pyramid fusion (FPF) integrates the features to obtain $\mathcal{H}_F^k$.
\begin{equation}
   \mathcal{FPF}({\mathcal H}_F^{k - 1},{{\mathcal I}_{1/({2^{k+1}})}},{\mathcal E}_F^k) \to {\mathcal H}_F^k
\end{equation}
\begin{equation}
\small
    \mathcal{FPF}({\mathcal H}_F^{k - 1},{{\mathcal I}_{1/({2^{k - 1}})}},{\mathcal E}_F^k) = \mathcal{H}_e({\mathcal H}_F^{k - 1} \oplus {{\mathcal I}_{1/({2^{k - 1}})}} \oplus {\mathcal E}_F^k)
\end{equation}
where $k=\{0,...,2\}$, \( \mathcal{H}_e \) consists of convolutional layers featuring a kernel size of 3 with a ReLU activation function. FPF has two advantages. First, it preserves the spatial structure of the image and features, allowing for the encoding of more local information. Second, it achieves fusion by simply concatenating features without additional convolutions or fully connected layers, improving efficiency.

The estimated edge details  $\mathcal{H}_r$ are learned through L1 loss, expressed as follows:
\begin{equation}
    L_\mathcal{H}=m_d\left\|\mathcal{H}_r-\alpha_g\right\|_1
\end{equation}
where $m_d$ is a binary boundary mask generated from $\alpha_g$ using dilation and erosion operations. When the pixel is located within the boundary transition region, the mask is set to 1; otherwise, it is set to 0.

\noindent\textbf{Cloth Cutting.} Clothing cutting integrates edge information and body semantic information to cut the clothing. Clothing cutting combines decoded features from edge prediction with those from the cutting process, enhancing the refinement of clothing cutting. We first use bilinear interpolation to upsample \(\mathcal{C}_F^k\) twofold, resulting in $\mathcal{C}_F^{{{k}^{'}}}$.
\begin{equation}
   {\mathcal{C}_F^{k'}}(x, y) = \sum_{i=0}^{1} \sum_{j=0}^{1} {\mathcal{C}_F^k}(2x-i, 2y-j) \cdot (i-x)_+ \cdot (j-y)_+
\end{equation}
where $ k =\{5,6\}$, and \( (i-x)_+ \) and \( (j-y)_+ \) are the weights defined as \( \max(i-x, 0) \) and \( \max(j-y, 0) \), ensuring that the weights are non-negative.
The process of Decoded Features Fusion (FDF) can be represented as follows:
\begin{equation}
   {\mathcal{FDF}}(\mathcal{H}_F^{k - 3},\mathcal{C}_F^{{{(k - 1)}^{'}}}) \to \mathcal{C}_F^k
\end{equation}
\begin{equation}
    {\mathcal{FDF}}(\mathcal{H}_F^{k - 3},\mathcal{C}_F^{{{(k - 1)}^{'}}})=\mathcal{C}_d(\mathcal{H}_F^{k - 3} \oplus\mathcal{C}_F^{{{(k - 1)}^{'}}})
\end{equation}
where $k=\{7,8\}$,$\mathcal{C}_d$ consists of convolutional layers with a kernel size of 1 followed by a ReLU activation layer. Through the FDF process, the edge stitching results are effectively fused with the clothing cutting features, enhancing the final cutting accuracy.

The clothing-cutting result $\mathcal{C}_r$ is predicted with the following constraints:
\begin{equation}
    {{L}_\mathcal{C} } = {\left\| {{{\mathcal C}_r} - {\alpha _g}} \right\|_1} + {{L}_{sl}}({{\mathcal C}_r},{\alpha _g})+ {{L}_{kl}}({{\mathcal C}_r},{\alpha _g})
\end{equation}
where \( L_{sl} \) is the similarity loss, and \( {L_{kl}} \) denotes the divergence loss.

Overall, the clothing cutting process is supervised by \(L_\mathcal{H}\), ${{L}_{cm}}$, and \({{L}_\mathcal{C}}\).
\begin{equation}
    L_{cloth} = \lambda_{h}L_\mathcal{H}+\lambda_{c}{{L}_\mathcal{C} } + \lambda_{m}L_{cm} 
\end{equation}
where \(\lambda_{c}=\lambda_{h}=4\) and \(\lambda_{m}=2\). By multiplying \(\mathcal{C}_r\) with the original image \(\mathcal{I}\), we obtain the human body \(\mathcal{B}\).


\begin{table*}[ht]

\caption{Quantitative comparisons with SOTA methods on four benchmark datasets.}
\label{tab:com}
\setlength{\tabcolsep}{2pt}
\begin{tabular}{l|l|c|ccc|ccc|ccc}
\hline
    \multirow{2}{*}{Methods} &\multirow{2}{*}{Publications}&\multicolumn{1}{c|}{ Cloth4D} &\multicolumn{3}{c|}{THuman2.0}&\multicolumn{3}{c|}{EMDB} &\multicolumn{3}{c}{ 3DPW} \\  \cline{3-12}  
    & & mIoU$\uparrow$& MPJPE$\downarrow$ & PA-MPJPE$\downarrow$ & MVPE$\downarrow$& MPJPE$\downarrow$ & PA-MPJPE$\downarrow$ & MVPE$\downarrow$& MPJPE$\downarrow$ &  PA-MPJPE$\downarrow$ & MVPE$\downarrow$  \\ 
\hline
 SPIN~\cite{SPIN}     &ICCV'2019  &0.706  & 91.3 & 65.5 & 114.0& 136.9 &81.7 & 168.7& 103.8 & 56.2 & 128.0	\\

PyMAF~\cite{pymaf}     &ICCV'2021 &0.724&87.5&63.0&108.7&128.7  & 79.8&158.3  &  97.4 & 54.9 & 119.1 \\
PyMAF-X~\cite{pymaf-X}  &TPAMI'2023 &0.766&84.2&58.2&96.0&  88.3&66.2 & 82.9 & 78.0  &  47.1 &  91.3 \\
VirtualMarker~\cite{Virtual_Markers}  &CVPR'2023&0.576&75.1 & 56.7& 92.0&  99.5&72.7 &104.6  &67.5& \textbf{41.3}&77.9\\

MeshPose~\cite{meshpose}&CVPR'2024&0.794&69.9&50.8&83.6&81.0&58.8&90.5&65.7&45.1&66.0\\
TokenHMR~\cite{tokenhmr} &CVPR'2024&0.799 &58.1&42.6&74.9&79.8 &57.2 &90.8  &60.8&43.7&62.8\\
 \hline
  ClothHMR &Ours	 & \textbf{0.846}&\textbf{46.9}&\textbf{35.6}&\textbf{63.6}& \textbf{71.7} &\textbf{50.6}&\textbf{73.6	}& \textbf{59.9}& 41.9& \textbf{59.2}   \\
\hline
\end{tabular}
\end{table*}
\subsection{Mesh Recovering }
\label{sec:MR}

Mesh recovery first estimates the initial human mesh \( \mathcal{M}_0 \) from \( \mathcal{B} \). Subsequently, it optimizes parameters by iterative fitting the intermediate representations of the human mesh with those obtained from the foundational human vision model, resulting in \( \mathcal{M}_r \).

\noindent\textbf{Foundational Human Vision Model.} Inspired by Sapiens~\cite{sapiens}, a cohesive encoder-decoder architecture is employed to fine-tune tasks related to human tasks. The encoder is initialized with weights pre-trained using MAE~\cite{MAE}, while the decoder is a lightweight network designed for various tasks and initialized randomly.

During pre-training, the image \( \mathcal{B} \) is first divided into fixed-size patches, denoted as \( \mathcal{B}_p \in \mathbb{R}^{\frac{H}{d} \times \frac{W}{d} \times C} \), where \( d \) is the number of patches, and \( C \) represents the number of channels. In the encoding phase, some of these patches are randomly masked, while the remaining unmasked patches are used as input. The encoder processes the visible patches, converting them into a latent representation. In the decoding phase, the decoder takes the output from the encoder, along with positional embeddings, to reconstruct the full set of image patches, including the previously masked ones, thereby restoring the complete image.

The joint detection task uses a lightweight decoder \(\mathcal{J}\) designed to convert the encoder’s output features into joint heatmaps. The decoder begins by mapping the encoder’s output \(F_{\text{out}}\) to a lower-dimensional feature space via a fully connected layer. Afterward, the decoder upsamples the features through two transposed convolution blocks, each composed of a transposed convolution layer, batch normalization, and ReLU activation. Each block doubles the size of the feature map. In the final step, a \(1 \times 1\) convolution layer generates the heatmaps for the joints. This process is mathematically expressed as:
\begin{equation}
    \mathcal{G}_{j} = \mathcal{Q}_{1 \times 1}(\mathcal{V}(\mathcal{V}(F_{\text{out}})))
\end{equation}
where $\mathcal{Q}$ represents convolution layers and $\mathcal{V}$ represents decoder. \(F_{\text{out}}\) refers to the encoder’s output. \(\mathcal{G}_{j} \in \mathbb{R}^{H \times W \times N_k}\) is the predicted heatmap for each joint, and \(N_k\) represents the number of joints to be predicted, which is set to 24. 

This process is trained to minimize the mean squared error (MSE) loss between the predicted heatmaps and the ground truth.
\begin{equation}
L_{\text{pose}} = \text{MSE}(J_{gt}, \mathcal{G}_{j})
\end{equation}
where \(J_{gt}\) represents the ground truth joint heatmaps.

The body depth estimation task employs a depth estimation decoder (\(\mathcal{D}\)) to regress the encoder's features $F_{\text{out}}$ into an output with a single channel. The depth map is given by $\mathcal{G}_{d} = \mathcal{D}(F_{\text{out}})$, where $\mathcal{G}_{d} \in \mathbb{R}^{{H} \times {W}}$. The depth values in \(\mathcal{G}_{d}\) are normalized to the range [0, 1], based on the maximum and minimum depths.
Similar to Sapiens, this process is trained to minimize the $L_{\text{depth}}$.
\begin{equation}
\begin{array}{c}
\Delta d = \log (D_{gt}) - \log ({\mathcal G}_{d})\\
{{L}_{{\rm{depth}}}} = \sqrt {\frac{1}{N_p}\sum\limits_{i = 1}^{N_p} {{{(\Delta d_i)}^2}}  - \frac{1}{2}{{(\frac{1}{N_p}\sum\limits_{i = 1}^{N_p} {(\Delta d_i)} )^2}}} 

\end{array}
\end{equation}
where \(D_{gt}\) represents the ground truth depth, $N_p$ represents the number of human pixels in the image.

The body silhouette extraction task employs a silhouette estimation decoder (\(\mathcal{M}\)), which is similar to \(\mathcal{D}\). The silhouette map is given by \(\mathcal{G}_{m} = \mathcal{M}(F_{\text{out}})\), where \(\mathcal{G}_{m} \in \mathbb{R}^{H \times W}\). This process is trained to minimize the $L_{\text{mask}}$.
\begin{equation}
    {{L}_{{mask}}} = ||{{M}_{gt}} - {{\mathcal G}_m}|{|_1}
\end{equation}

\noindent\textbf{Iterative Fitting.}
After extracting the intermediate representation from the input image through FHVM, we employ a simple iterative fitting strategy to recover 3D human meshes. In ClothHMR, the 3D human meshes are represented using the SMPL model~\cite{smpl}. The depth map \(\mathcal{P}_d\) of the SMPL model can be generated by calculating the distances between the model points. The mask map \(\mathcal{P}_m\) and the 2D joints \(\mathcal{P}_j\) of the SMPL model can be obtained by rendering the model onto the 2D plane using the camera parameters. By aligning these intermediate representations, we optimize the human mesh generation.
To initialize the process, we first estimate an initial mesh \(\mathcal{M}_0\) from the tailored image \(\mathcal{B}\) using methods such as   Subsequently, the parameters of the mesh are refined through iterative fitting for greater accuracy.
We optimize SMPL's shape, $\beta$, pose, $\theta$, scale, $s$ and translation, $t$, to minimize:  
\begin{equation}
\begin{array}{l}
    {{L}_{smpl}} = {\lambda _{\rm{d}}}{{L}_D} + {\lambda _m}{{L}_m} + {\lambda _j}{{L}_J}\\
{{L}_{{D}}} = ||{{\mathcal G}_d} - {{\mathcal P}_d}|{|_1}\\
{{L}_{{M}}} = ||{{\mathcal G}_m} - {{\mathcal P}_m}|{|_1}\\
{{L}_{{J}}} = ||{{\mathcal G}_j} - {{\mathcal P}_j}|{|_2}
\end{array}
\end{equation}
where $\lambda _{{d}}$, $\lambda _{{m}}$, and $\lambda _{{j}}$ are three balancing weights, with \(\lambda_d = 5\), \(\lambda_m = 5\), and \(\lambda_j = 10\). \(\mathcal{G}_d\), \(\mathcal{G}_m\), and \(\mathcal{G}_j\) are obtained from FHVM.

  
\begin{figure*}[h]
  \includegraphics[width=0.85\textwidth]{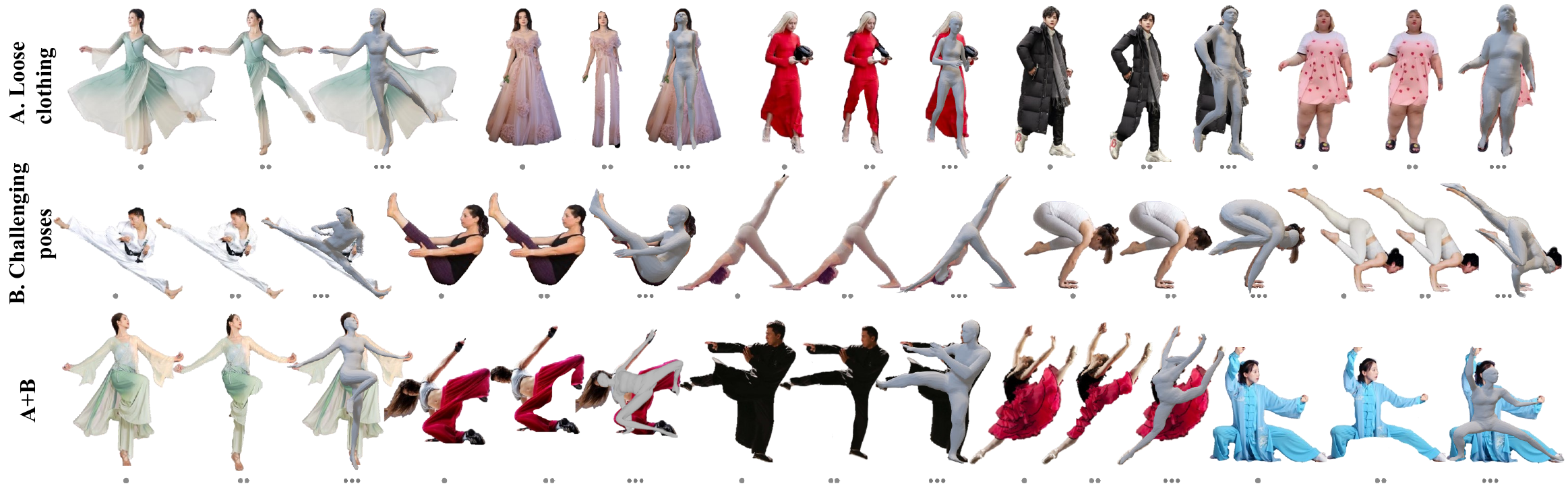}
  \caption{Results produced by ClothHMR.  For each input image ($\cdot$), we show the intermediate cloth tailoring result ($\cdot\cdot$) and the final 3D human mesh recovery result ($\cdot\cdot\cdot$). ClothHMR can accurately recover 3D human meshes while addressing the challenges posed by various loose clothing and complex poses. Please zoom in to see the details. }
  
  \label{Fig:rmcloth}
\end{figure*}

\section{Experimental Results}

\subsection{Implementation}

When training the CT module, we set the batch size to 16 and the initial learning rate to 0.01. The learning rate is decreased by a factor of 0.1 every 10 epochs. We train the CT module for 60 epochs, and the process takes about 10 days on a single V100 GPU.
We construct the foundational human vision model using the pre-trained Sapiens~\cite{sapiens} model and fine-tune it for three tasks: joint detection, depth estimation, and silhouette extraction. All tasks utilize the AdamW optimizer and are trained for approximately 25 days on 16 V100 GPUs.
In the iterative fitting of MR, the number of optimization iterations is set to 40.

\subsection{Datasets and Metrics}

\textbf{Training Datasets. }
We combine the synthetic 3D human dataset Cloth4D~\cite{cloth4d}  and the scanned 3D human dataset THuman2.0~\cite{THuman2.0} for training the CT module. The Cloth4D dataset contains 100,000 synthetic 3D human models of 1,000 virtual subjects with varied poses and clothing, and THuman2.0 comprises 525 scanned 3D human models of college students with daily clothing. We use 80\% of the data from both datasets for training, reserving the remaining data for testing.
Training the CT module requires paired images of clothed and unclothed humans. Cloth4D officially provides 8 pairs of rendered images for each 3D clothed human model and its SMPL model, which can be directly used to train the CT module. While THuman2.0 provides 3D clothed human meshes and SMPL models, we render 18 paired images from each 3D clothed human and its SMPL model.

We use Cloth4D~\cite{cloth4d} and THuman2.0~\cite{THuman2.0} to fine-tune the FHVM. To fine-tune FHVM, the required data includes images, 2D joints, 2D silhouettes, and depth information. THuman2.0 renders mesh from 8 views to obtain 1024x1024 images, and 2D joints are derived from the joints of the SMPL model. The 2D silhouettes are extracted from the images, and depth information is calculated from the 3D mesh. Cloth4D includes non-standard SMPL models, so the 2D joints are obtained through 
downsampled SMPL models. Other attribute extractions are consistent with those in THuman2.0.

\noindent\textbf{Testing Datasets.}
To extensively evaluate ClothHMR, in addition to assessments on the Cloth4D and THuman2.0 testing sets, we conduct cross-dataset evaluations on the EMDB~\cite{EMDB} and 3DPW~\cite{3dpw} benchmark datasets as well as in-the-wild images collected from a web app. EMDB contains 81 indoor and outdoor videos of 10 subjects with unconstrained poses and daily clothing. 
We extract 24,114 images from these videos for testing, which is identical to the experimental setup of TokenHMR~\cite{tokenhmr}.
3DPW, which is commonly used in evaluating 3D human mesh recovery methods, is composed of 60 in-the-wild videos of humans with complex poses. 
In our experiments, we directly use the 3DPW testing set, which contains 11,376 images.

\noindent\textbf{Metrics.}
Since the Cloth4D dataset only provides non-standard SMPL models, we adopt mIoU (mean Intersection over Union) as the evaluation metric. The THuman2.0, EMDB, and 3DPW datasets provide annotated SMPL models, so we report three commonly used metrics: MPJPE (Mean Per Joint Position Error), PA-MPJPE (Procrustes Aligned MPJPE), and MVPE (Mean Vertex Position Error).

\subsection{Quantitative Evaluation}

\begin{figure}[ht]
  \includegraphics[width=\linewidth]{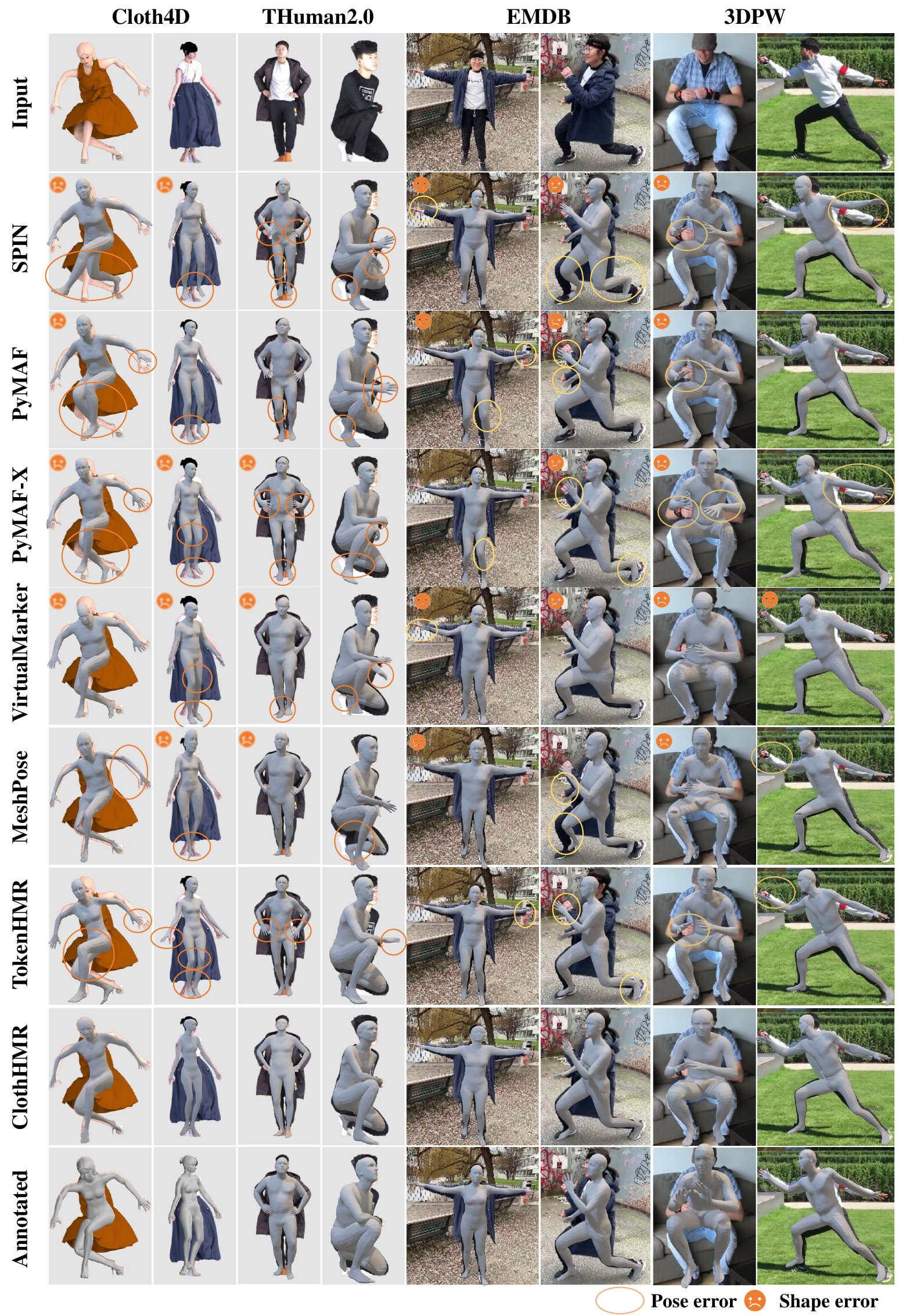}
  \caption{Visual comparison with other SOTA methods on the Cloth4D, THuman2.0, EMDB and 3DPW datasets. Please zoom in to see the details. }
  \Description{Visual comparison with other SOTA methods on Cloth4D, THuman2.0, EMDB and 3DPW datasets.}
  \label{Fig:com_dataset}
\end{figure}

ClothHMR is compared with five SOTA single-image 3D human mesh recovery methods with publicly released code and models: SPIN~\cite{SPIN}, PyMAF~\cite{pymaf}, PyMAF-X~\cite{pymaf-X}, VirtualMarker~\cite{Virtual_Markers}, MeshPose~\cite{meshpose} and TokenHMR~\cite{tokenhmr}. 
To ensure optimal performance, we use the original code and trained models provided by the authors of these methods.
As shown in Table~\ref{tab:com}, ClothHMR consistently demonstrates superior performance and surpasses the other comparative methods across all four benchmark datasets.
Compared to the current SOTA method, transformer-based TokenHMR~\cite{tokenhmr}, ClothHMR demonstrates a 5.9\% improvement in mIoU on the Cloth4D dataset. Meanwhile, ClothHMR achieves reductions of 19.3\%, 16.4\%, and 15.1\% in MPJPE, PA-MPJPE, and MVPE on the THuman2.0 dataset; while on the EMDB dataset, it reduces these metrics by 10.2\%, 11.5\%, and 18.9\%, respectively.
Despite the prevalence of tight clothing worn by subjects in the 3DPW dataset, ClothHMR still achieves reductions of 1.5\%, 4.1\%, and 5.7\% in MPJPE, PA-MPJPE, and MVPE, respectively, compared to TokenHMR.

\subsection{Qualitative Evaluation}

\textbf{Visual Results.}
Fig.~\ref{Fig:rmcloth} illustrates the results produced by ClothHMR on in-the-wild images. These images feature a diverse range of body types for both men and women, wearing various types of clothing such as casual wear, performance costumes, and sportswear, while posing in complex actions. The results show that ClothHMR successfully recovers 3D meshes that accurately capture the subjects' body shapes and poses.

\begin{figure}[ht]
  \includegraphics[width=\linewidth]{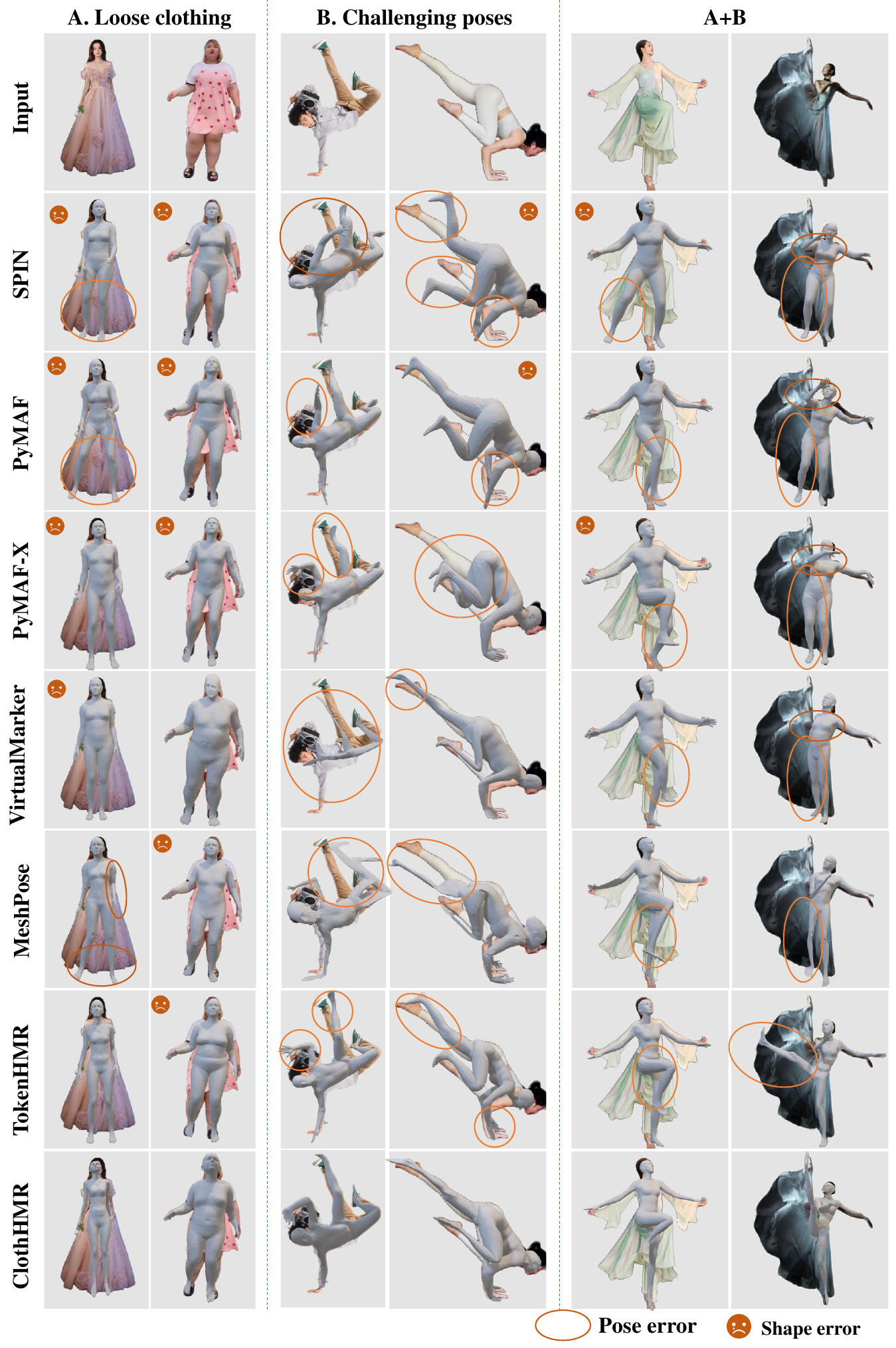}
  \caption{Visual comparison with other SOTA methods on in-the-wild images. Please zoom in to see the details. }
  \Description{Visual comparison with other SOTA methods on Cloth4D, THuman2.0, EMDB and 3DPW datasets.}
  \label{Fig:com_real_data}
\end{figure}

\begin{table*}[h]

    \caption{Ablation comparisons on the Cloth4D, THuman2.0, EMDB test set.}
    \label{tab:abl}
    \begin{tabular}{c|c|ccc|ccc}
\hline
    \multirow{2}{*}{Variants} &\multicolumn{1}{c|}{ Cloth4D} &\multicolumn{3}{c|}{THuman2.0}&\multicolumn{3}{c}{EMDB} \\ \cline{2-8}
    & mIoU$\uparrow$& MPJPE$\downarrow$ & PA-MPJPE$\downarrow$ & MVPE$\downarrow$& MPJPE$\downarrow$ & PA-MPJPE$\downarrow$& MVPE$\downarrow$  \\
\hline

 ClothHMR (without CT) & 0.768&  68.5&48.6&85.5&87.4&60.2&90.9\\
  CT+TokenHMR& 0.806& 56.1&41.4&71.6&73.3&53.7&83.8 \\
  ClothHMR (without FHVM)& 0.809& 61.4&43.8&80.6&78.7&55.3&85.7 \\
  ClothHMR (withFHVM)   	 & 	0.846 &46.9&35.6&63.6&71.7&50.6&73.6  \\
\hline
\end{tabular}
\end{table*}
\noindent\textbf{Qualitative Comparisons.}
Fig.~\ref{Fig:com_dataset} presents a visual comparison of ClothHMR with other SOTA methods
across the Cloth4D, THuman2.0, EMDB, and 3DPW benchmark datasets.
It can be observed that most other methods exhibit incorrect human pose estimation on the Cloth4D, THuman2.0, and EMDB datasets due to the occlusion caused by loose clothing. Moreover, SPIN, PyMAF-X, VirtualMarker, and MeshPose tend to estimate body shape inaccurately. In contrast, ClothHMR demonstrates the ability to estimate both human pose and body shape accurately across various types of clothing.
On the 3DPW dataset, where subjects are not wearing loose clothing, ClothHMR also outperforms the other methods.

Fig.~\ref{Fig:com_real_data} compares ClothHMR and other SOTA methods on in-the-wild images. The individuals in these images are either dressed in very loose clothing or exhibit complex poses, both of which make the recovery of their 3D meshes particularly challenging. The results show that other methods struggle with these challenges and may produce extremely wrong 3D pose estimates. While TokenHMR can estimate approximate poses, their predictions lack precision. Moreover, none of these methods can accurately estimate body shapes concealed beneath clothing. In contrast, ClothHMR consistently delivers superior performance on these in-the-wild images.

\begin{figure}[h]
  \centering
  \includegraphics[width=0.9\linewidth]{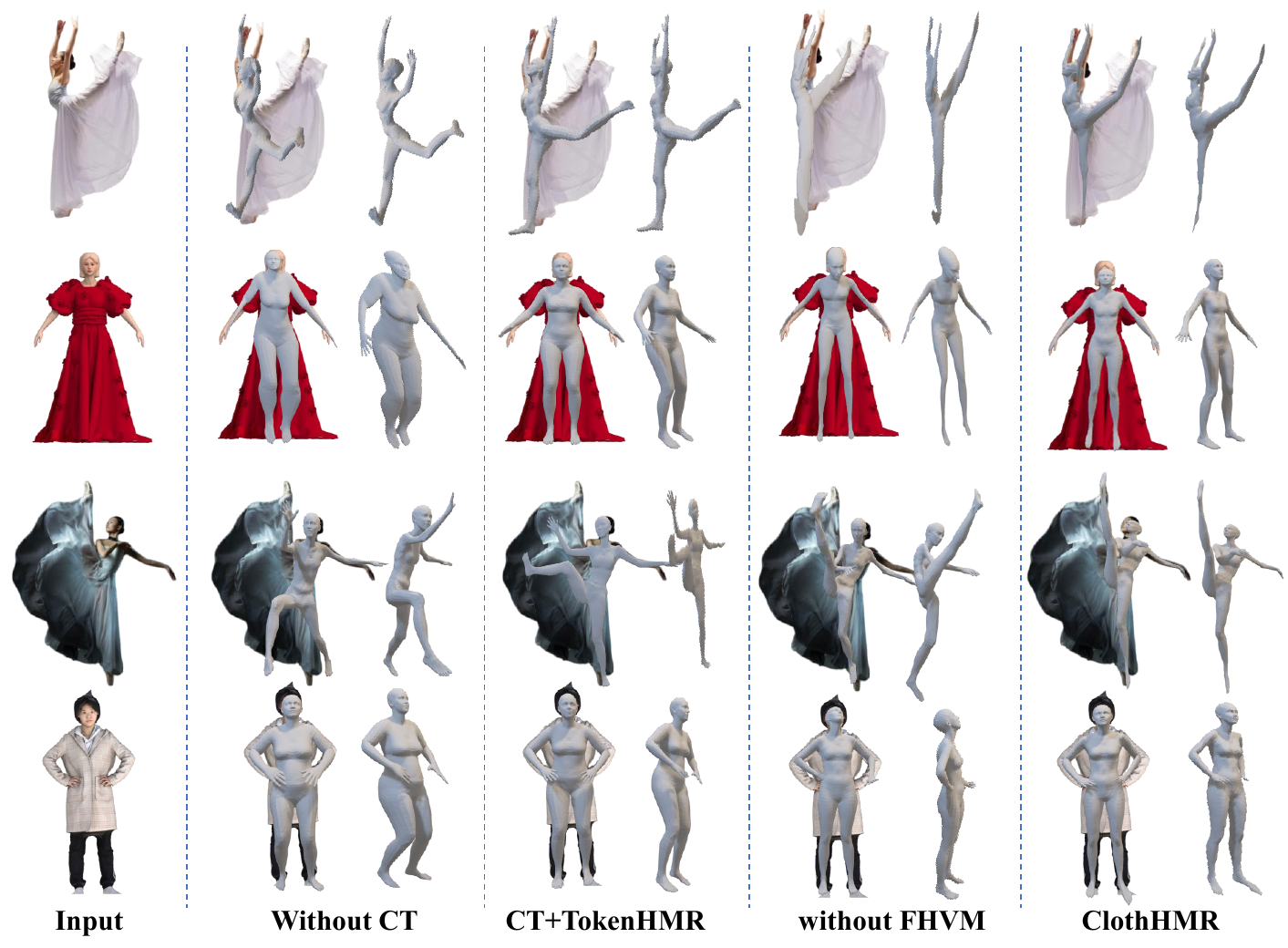}
  \caption{ Results produced by different variants.}
  \Description{}
  \label{Fig:abl}
\end{figure}
\subsection{Ablation Study}

\textbf{Effectiveness of Cloth Tailoring.}
To evaluate the effectiveness of the CT module, we implement a variant with the CT module removed. As reported in Table~\ref{tab:abl}, incorporating the CT module contributes to a 10.2\% improvement in mIoU on Cloth4D. Additionally, it reduces MPJPE, PA-MPJPE, and MVPE on THuman2.0 and EMDB by 31.5\%, 24.7\%, and 25.6\%, and by 18.0\%, 15.9\%, and 19.0\%, respectively. These results indicate that the clothing tailoring scheme can significantly mitigate the impact of clothing on 3D mesh human recovery. 
The visual results in Fig.~\ref{Fig:abl} show that the removal of the CT module may lead to significant pose and body shape errors, reinforcing the conclusions drawn from Table~\ref{tab:abl}.

\noindent\textbf{Performance of the Mesh Recovering Module.}
We replace the MR module in ClothHMR with the current SOTA TokenHMR~\cite{tokenhmr}. Compared to TokenHMR, our MR module achieves a 5.0\% enhancement in mIoU on Cloth4D, and reduces MPJPE, PA-MPJPE, and MVPE on THuman2.0 and EMDB by 16.4\%, 14.0\%, and 11.2\%, and by 2.2\%, 5.8\%, and 12.2\%, respectively.  

\noindent\textbf{Effectiveness of Foundational Human Vision Model.}
To evaluate the contribution of FHVM, we replace it with a combination of OpenPose~\cite{openpose}, MODNet~\cite{MODNet}, and HDNet~\cite{hdnet}. 
The results presented in Table~\ref{tab:abl}
indicate that integrating the FHVM enhances the mIoU on Cloth4D by 4.6\%. Additionally, it significantly reduces MPJPE, PA-MPJPE, and MVPE on THuman2.0 and EMDB by 23.6\%, 18.7\%, and 21.1\%, and by 8.9\%, 8.5\%, and 14.2\%, respectively.

\section{Web Applications}
\begin{figure}[h]
  \centering
  \includegraphics[width=\linewidth]{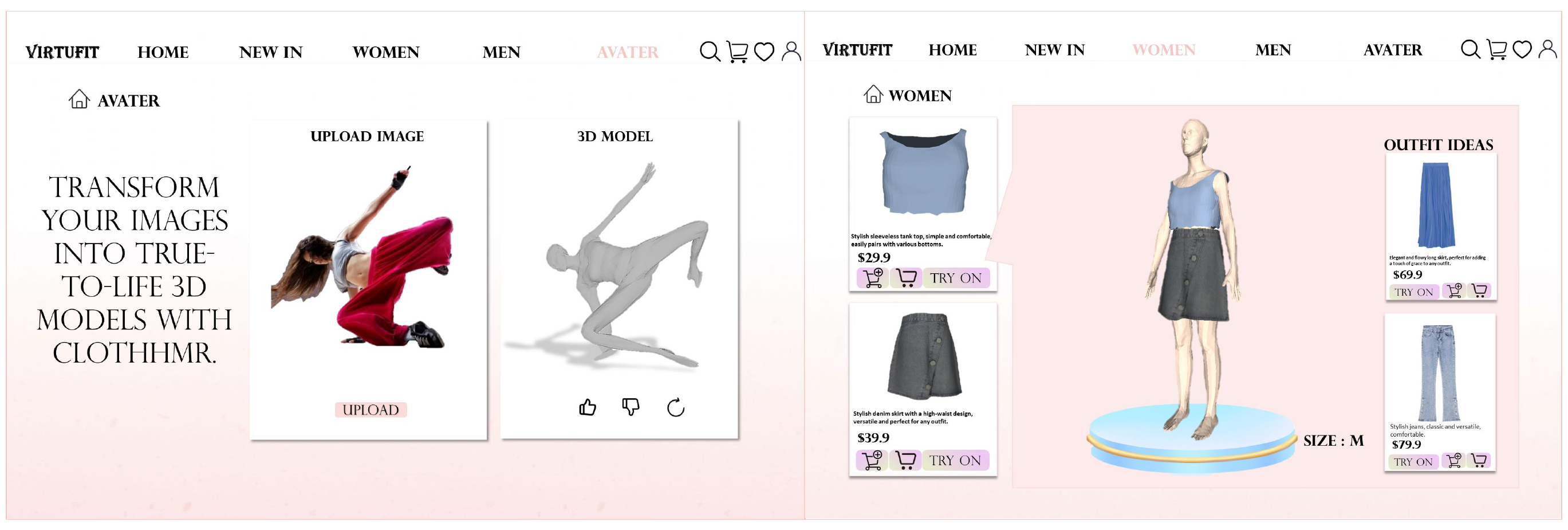}
  \caption{Web application based on ClothHMR. ClothHMR can accurately estimate users' body shapes and poses in real-world usage scenarios. The top displays the webpage where users create their models, while the bottom shows the shopping virtual fitting page.
 }
  \Description{}
  \label{Fig:web}
\end{figure}

As shown in Fig.~\ref{Fig:web}, we developed a web application for online fashion and shopping based on ClothHMR.
For a virtual try-on application, only if it can accurately estimate the users' 3D meshes, regardless of the type of clothing they are currently wearing, can it provide a seamless user experience. Existing 3D mesh recovery methods struggle to achieve this, whereas ClothHMR can accurately estimate users' body shapes and poses in real-world usage scenarios. 
In the future, we will develop more web applications based on ClothHMR, including online fitness, personalized content creation, and immersive gaming experiences.

\section{Conclusion}
We propose ClothHMR, a method for accurately recovering 3D human mesh models from various clothing and poses. The innovations of ClothHMR are primarily divided into two modules: the clothing tailoring module (CT) and the mesh recovering module (MR). Given an input image, the CT module trims the excess parts of the clothing through body semantic estimation and edge prediction. An initial approach is then used to estimate the initial 3D human mesh. The MR module aligns the intermediate representations from the foundational human vision model (FHVM) with those from the mesh through iterative optimization, ultimately generating an accurate human mesh. Experiment results demonstrate that ClothHMR can accurately recover 3D meshes of individuals wearing a variety of clothing and precisely estimate their body shapes and poses. Across four benchmark datasets and in-the-wild images, ClothHMR significantly outperforms existing SOTA methods. In addition, a web application for online fashion and shopping, powered by ClothHMR, has been developed, demonstrating ClothHMR's effectiveness in real-world applications.

\begin{figure}[h]
  \centering
  \includegraphics[width=0.6\linewidth]{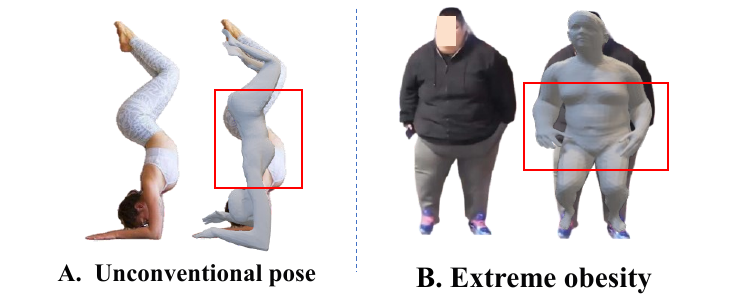}
  \caption{Failure cases. ClothHMR may fail to estimate accurate poses and body shapes when handling unconventional poses and extreme obesity. }
  \Description{}
  \label{Fig:failure_case}
\end{figure}

\textbf{Limitations \& Future Work.}
Although ClothHMR achieves impressive results, it still has some limitations when handling unconventional poses and extreme obesity, as shown in Fig.~\ref{Fig:failure_case}.
The reason behind this is that ClothHMR employs the parameterized SMPL model, and the human body in these scenarios exceeds the expressive potential of SMPL~\cite{smpl}. In future work, we will adopt non-parameterized models~\cite{VS} to represent 3D human meshes to overcome these limitations. 

\section{Acknowledgments}
This work was supported by the National Natural Science Foundation of China (62077026), and the Fundamental Research Funds for the Central Universities (CCNU22QN012).

\bibliographystyle{ACM-Reference-Format}
\balance
\bibliography{sample-base}

@String{Computer = "{IEEE} Computer" }

@String{Springer = "Springer-Verlag" }

@String{PAMI = "IEEE Trans. Pattern Anal. Mach. Intell. (TPAMI)"}

@String{CVPR= "Conf. Comput. Vis. Pattern Recog. (CVPR)"}

@String{ICCV= "Int. Conf. Comput. Vis. (ICCV)"}

@String{ECCV= "Eur. Conf. Comput. Vis. (ECCV)"}

@String{NIPS= "Adv. Neural Inform. Process. Syst. (NeurIPS)"}

@String{BMVC= "Brit. Mach. Vis. Conf. (BMVC)"}

@String{TOG= "ACM Trans. Graph. (TOG)"}

@String{ICLR = "Int. Conf. Learn. Represent."}

@String{AAAI = "AAAI"}

@INPROCEEDINGS{icon,
  author={Xiu, Yuliang and Yang, Jinlong and Tzionas, Dimitrios and Black, Michael J.},
  booktitle=CVPR, 
  title={ICON: Implicit Clothed humans Obtained from Normals}, 
  year={2022},
  volume={},
  number={},
  pages={13286-13296},
 
  doi={10.1109/CVPR52688.2022.01294}}

@INPROCEEDINGS{SLVM,
  author={Bai, Yutong and Geng, Xinyang and Mangalam, Karttikeya and Bar, Amir and Yuille, Alan L. and Darrell, Trevor and Malik, Jitendra and Efros, Alexei A.},
  booktitle=CVPR, 
  title={Sequential Modeling Enables Scalable Learning for Large Vision Models}, 
  year={2024},
  volume={},
  number={},
  pages={22861-22872},
  doi={10.1109/CVPR52733.2024.02157}}

@INPROCEEDINGS{MAE,
  author={He, Kaiming and Chen, Xinlei and Xie, Saining and Li, Yanghao and Dollár, Piotr and Girshick, Ross},
  booktitle=CVPR, 
  title={Masked Autoencoders Are Scalable Vision Learners}, 
  year={2022},
  volume={},
  number={},
  pages={15979-15988},
  doi={10.1109/CVPR52688.2022.01553}}

@inproceedings{VIT,
title	= {An Image is Worth 16x16 Words: Transformers for Image Recognition at Scale},
author	= {Alexander Kolesnikov and Alexey Dosovitskiy and Dirk Weissenborn and Georg Heigold and Jakob Uszkoreit and Lucas Beyer and Matthias Minderer and Mostafa Dehghani and Neil Houlsby and Sylvain Gelly and Thomas Unterthiner and Xiaohua Zhai},
booktitle=ICLR,
year	= {2021}}

@INPROCEEDINGS{MAE1,
  author={Singh, Mannat and Duval, Quentin and Alwala, Kalyan Vasudev and Fan, Haoqi and Aggarwal, Vaibhav and Adcock, Aaron and Joulin, Armand and Dollár, Piotr and Feichtenhofer, Christoph and Girshick, Ross and Girdhar, Rohit and Misra, Ishan},
  booktitle=ICCV, 
  title={The effectiveness of MAE pre-pretraining for billion-scale pretraining}, 
  year={2023},
  volume={},
  number={},
  pages={5461-5471},
  doi={10.1109/ICCV51070.2023.00505}}

@INPROCEEDINGS{meshpose,
  author={Le, Eric-Tuan and Kakolvris, Antonis and Koutras, Petros and Tam, Himmy and Skordos, Efstratios and Papandreou, George and Güler, Riza Alp and Kokkinos, Iasonas},
  booktitle={IEEE/CVF Conference on Computer Vision and Pattern Recognition (CVPR)}, 
  title={{MeshPose}: Unifying DensePose and 3D Body Mesh reconstruction}, 
  year={2024},
  volume={},
  number={},
  pages={2405-2414},
  doi={10.1109/CVPR52733.2024.00233}}

@article{SMPL,
  title={SMPL: A skinned multi-person linear model},
  author={Loper, Matthew and Mahmood, Naureen and Romero, Javier and Pons-Moll, Gerard and Black, Michael J},
  journal=TOG,
  volume={34},
  number={6},
  pages={1-16},
  year={2015},
  doi={http://doi.org/10.1145/2816795.2818013}
}

@inproceedings{ProHMR,
  author={Kolotouros, Nikos and Pavlakos, Georgios and Jayaraman, Dinesh and Daniilidis, Kostas},
  booktitle=ICCV, 
  title={Probabilistic modeling for human mesh recovery}, 
  year={2021},
  volume={},
  number={},
  pages={11585-11594},
  doi={http://doi.org/10.1109/ICCV48922.2021.01140}}

@inproceedings{DenseRaC,
  author={Xu, Yuanlu and Zhu, Song-Chun and Tung, Tony},
  booktitle=ICCV, 
  title={{DenseRaC}: Joint {3D} Pose and Shape Estimation by Dense Render-and-Compare}, 
  year={2019},
  volume={},
  number={},
  pages={7759-7769},
  doi={http://doi.org/10.1109/ICCV.2019.00785}}

@inproceedings{zanfir2020weakly,
  title={Weakly supervised {3D} human pose and shape reconstruction with normalizing flows},
  author={Zanfir, Andrei and Bazavan, Eduard Gabriel and Xu, Hongyi and Freeman, William T and Sukthankar, Rahul and Sminchisescu, Cristian},
  booktitle=ECCV,
  pages={465-481},
  year={2020},
  organization={Springer},
  doi = {http://doi.org/10.1007/978-3-030-58539-6_28},
}

@inproceedings{HybrIK,
  author={Li, Jiefeng and Xu, Chao and Chen, Zhicun and Bian, Siyuan and Yang, Lixin and Lu, Cewu},
  booktitle=CVPR, 
  title={{HybrIK}: A Hybrid Analytical-Neural Inverse Kinematics Solution for {3D} Human Pose and Shape Estimation}, 
  year={2021},
  volume={},
  number={},
  pages={3382-3392},
  doi={http://doi.org/10.1109/CVPR46437.2021.00339}}

@inproceedings{chained,
  author={Nadine Rueegg and Christoph Lassner and Michael J. Black and Konrad Schindler},
  title={Chained Representation Cycling: Learning to Estimate 3D Human Pose and Shape by Cycling Between Representations},
  year={2020},
  cdate={1577836800000},
  pages={5561-5569},
  url={https://doi.org/10.1609/aaai.v34i04.6008},
  booktitle=AAAI,
}

@INPROCEEDINGS{VS,
  author={Liu, Leyuan and Li, Yuhan and Gao, Yunqi and Gao, Changxin and Liu, Yuanyuan and Chen, Jingying},
  booktitle=CVPR, 
  title={VS: Reconstructing Clothed 3D Human from Single Image via Vertex Shift}, 
  year={2024},
  volume={},
  number={},
  pages={10498-10507},
  
  doi={10.1109/CVPR52733.2024.00999}}

@inproceedings{DSR,
  title={Learning to regress bodies from images using differentiable semantic rendering},
  author={Dwivedi, Sai Kumar and Athanasiou, Nikos and Kocabas, Muhammed and Black, Michael J},
  booktitle=ICCV,
  pages={11250-11259},
  year={2021},
doi={http://doi.org/10.1109/ICCV48922.2021.01106}
}

@inbook{Sim2real,
author = {Doersch, Carl and Zisserman, Andrew},
title = {Sim2real transfer learning for 3D human pose estimation: motion to the rescue},
year = {2019},
publisher = {Curran Associates Inc.},
address = {Red Hook, NY, USA},

booktitle = NIPS,
articleno = {1160},
numpages = {13}
}

@article{PaMIR,
  title={{PaMIR}: Parametric model-conditioned implicit representation for image-based human reconstruction},
  author={Zheng, Zerong and Yu, Tao and Liu, Yebin and Dai, Qionghai},
  journal=PAMI,
  volume={44},
  number={6},
  pages={3170-3184},
  year={2021},
  doi={http://doi.org/10.1109/TPAMI.2021.3050505}
}

@inproceedings{HMR,
  title={End-to-end recovery of human shape and pose},
  author={Kanazawa, Angjoo and Black, Michael J and Jacobs, David W and Malik, Jitendra},
  booktitle=CVPR,
  pages={7122-7131},
  year={2018},
  doi={http://doi.org/10.1109/CVPR.2018.00744}
}

@inproceedings{skeleton2mesh,
  title={{Skeleton2Mesh}: Kinematics prior injected unsupervised human mesh recovery},
  author={Yu, Zhenbo and Wang, Junjie and Xu, Jingwei and Ni, Bingbing and Zhao, Chenglong and Wang, Minsi and Zhang, Wenjun},
  booktitle=ICCV,
  pages={8619-8629},
  year={2021},
  doi={http://doi.org/10.1109/ICCV48922.2021.00850}
}

@INPROCEEDINGS{SMPL-X,
  author={Pavlakos, Georgios and Choutas, Vasileios and Ghorbani, Nima and Bolkart, Timo and Osman, Ahmed A. and Tzionas, Dimitrios and Black, Michael J.},
  booktitle=CVPR, 
  title={Expressive Body Capture: 3D Hands, Face, and Body From a Single Image}, 
  year={2019},
  volume={},
  number={},
  pages={10967-10977},
  
  doi={10.1109/CVPR.2019.01123}}

@inproceedings{tung,
  title={Self-supervised learning of motion capture},
  author={Tung, Hsiao-Yu and Tung, Hsiao-Wei and Yumer, Ersin and Fragkiadaki, Katerina},
  booktitle=NIPS,
  pages = {5242–5252},
  year={2017},
http={https://dlnext.acm.org/doi/abs/10.5555/3295222.3295276}
}

@INPROCEEDINGS{neural,
  author={Zanfir, Andrei and Bazavan, Eduard Gabriel and Zanfir, Mihai and Freeman, William T. and Sukthankar, Rahul and Sminchisescu, Cristian},
  booktitle=CVPR, 
  title={Neural Descent for Visual 3D Human Pose and Shape}, 
  year={2021},
  volume={},
  number={},
  pages={14479-14488},
  doi={10.1109/CVPR46437.2021.01425}}

@inproceedings{SMPLify,
  title={{Keep it SMPL}: Automatic estimation of {3D} human pose and shape from a single image},
  author={Bogo, Federica and Kanazawa, Angjoo and Lassner, Christoph and Gehler, Peter and Romero, Javier and Black, Michael J},
  booktitle=ECCV,
  pages={561-578},
  year={2016},
  doi={http://doi.org/10.1007/978-3-319-46454-1\_34}
}

@inproceedings{GCMR,
  title={Convolutional mesh regression for single-image human shape reconstruction},
  author={Kolotouros, Nikos and Pavlakos, Georgios and Daniilidis, Kostas},
  booktitle=CVPR,
  pages={4501-4510},
  year={2019},
  doi={http://doi.org/10.1109/CVPR.2019.00463}
}

@inproceedings{SPIN,
  title={Learning to reconstruct {3D} human pose and shape via model-fitting in the loop},
  author={Kolotouros, Nikos and Pavlakos, Georgios and Black, Michael J and Daniilidis, Kostas},
  booktitle=ICCV,
  pages={2252-2261},
  year={2019},
  doi={http://doi.org/10.1109/ICCV.2019.00234}
}

@inproceedings{DecoMR,
  title={{3D} human mesh regression with dense correspondence},
  author={Zeng, Wang and Ouyang, Wanli and Luo, Ping and Liu, Wentao and Wang, Xiaogang},
  booktitle=CVPR,
  pages={7054-7063},
  year={2020},
  doi={http://doi.org/10.1109/CVPR42600.2020.00708}
}

@inproceedings{THUman2.0,
  title={{Function4D}: Real-time human volumetric capture from very sparse consumer rgbd sensors},
  author={Yu, Tao and Zheng, Zerong and Guo, Kaiwen and Liu, Pengpeng and Dai, Qionghai and Liu, Yebin},
  booktitle=CVPR,
  pages={5746-5756},
  year={2021},
   doi={http://doi.org/10.1109/CVPR46437.2021.00569}
}

@article{Human3.6M,
  title={{Human3.6M}: Large scale datasets and predictive methods for {3D} human sensing in natural environments},
  author={Ionescu, Catalin and Papava, Dragos and Olaru, Vlad and Sminchisescu, Cristian},
  journal=PAMI,
  volume={36},
  number={7},
  pages={1325-1339},
   doi={http://doi.org/10.1109/TPAMI.2013.248},
  year={2013}
}

@inproceedings{ HigherHRNet,
  author={Cheng, Bowen and Xiao, Bin and Wang, Jingdong and Shi, Honghui and Huang, Thomas S. and Zhang, Lei},
  booktitle=CVPR, 
  title={{HigherHRNet}: Scale-Aware Representation Learning for Bottom-Up Human Pose Estimation}, 
  year={2020},
  volume={},
  number={},
  pages={5385-5394},
  doi={http://doi.org/10.1109/CVPR42600.2020.00543}}

@inproceedings{LSP,
  title={Clustered Pose and Nonlinear Appearance Models for Human Pose Estimation},
  author={Johnson, Sam and Everingham, Mark},
  booktitle=BMVC,
  volume={2},
  number={4},
  pages={5},
  year={2010},
  doi = {http://doi.org/10.5244/C.24.12}
}

@article{SCAPE,
  title={{SCAPE}: Shape Completion and Animation of People},
  author={Anguelov, Dragomir and Srinivasan, Praveen and Koller, Daphne and Thrun, Sebastian and Rodgers, Jim and Davis, James},
  journal = TOG,
  volume = {24},
  number = {3},
  pages={408-416},
  year={2005},
  doi = {http://doi.org/10.1145/1073204.1073207},
}

@inproceedings{Pose2Mesh,
  title={{Pose2Mesh}: Graph convolutional network for {3D} human pose and mesh recovery from a {2D} human pose},
  author={Choi, Hongsuk and Moon, Gyeongsik and Lee, Kyoung Mu},
  booktitle=ECCV,
  pages={769-787},
  doi = {http://doi.org/10.1007/978-3-030-58571-6\_45},
  year={2020}
}

@inproceedings{Unite_the_People,
  title={{Unite the People}: Closing the loop between {3D} and {2D} human representations},
  author={Lassner, Christoph and Romero, Javier and Kiefel, Martin and Bogo, Federica and Black, Michael J and Gehler, Peter V},
  booktitle=CVPR,
  pages={6050-6059},
  doi={http://doi.org/10.1109/CVPR.2017.500},
  year={2017}
}

@INPROCEEDINGS{deepPose,
  author={Toshev, Alexander and Szegedy, Christian},
  booktitle=CVPR, 
  title={DeepPose: Human Pose Estimation via Deep Neural Networks}, 
  year={2014},
  volume={},
  number={},
  pages={1653-1660},
  
  doi={10.1109/CVPR.2014.214}}

@ARTICLE{openpose,
  author={Cao, Zhe and Hidalgo, Gines and Simon, Tomas and Wei, Shih-En and Sheikh, Yaser},
  journal=PAMI, 
  title={OpenPose: Realtime Multi-Person 2D Pose Estimation Using Part Affinity Fields}, 
  year={2021},
  volume={43},
  number={1},
  pages={172-186},
  doi={10.1109/TPAMI.2019.2929257}}

@ARTICLE{alphapose,
  author={Fang, Hao-Shu and Li, Jiefeng and Tang, Hongyang and Xu, Chao and Zhu, Haoyi and Xiu, Yuliang and Li, Yong-Lu and Lu, Cewu},
  journal=PAMI, 
  title={AlphaPose: Whole-Body Regional Multi-Person Pose Estimation and Tracking in Real-Time}, 
  year={2023},
  volume={45},
  number={6},
  pages={7157-7173},
  doi={10.1109/TPAMI.2022.3222784}}

@INPROCEEDINGS{Zanfir,
  author={Zanfir, Andrei and Marinoiu, Elisabeta and Sminchisescu, Cristian},
  booktitle=CVPR, 
  title={Monocular 3D Pose and Shape Estimation of Multiple People in Natural Scenes: The Importance of Multiple Scene Constraints}, 
  year={2018},
  volume={},
  number={},
  pages={2148-2157},
  doi={10.1109/CVPR.2018.00229}}

@inproceedings{deeplabv3+,
  title={Encoder-Decoder with Atrous Separable Convolution for Semantic Image Segmentation},
  author={Liang-Chieh Chen and Yukun Zhu and George Papandreou and Florian Schroff and Hartwig Adam},
  booktitle=ECCV,
  year={2018},
doi={https://doi.org/10.1007/978-3-030-01234-2_49}
}

@InProceedings{MODNet,
  author = {Zhanghan Ke and Jiayu Sun and Kaican Li and Qiong Yan and Rynson W.H. Lau},
  title = {MODNet: Real-Time Trimap-Free Portrait Matting via Objective Decomposition},
  booktitle = AAAI,
  year = {2022},
doi={https://doi.org/10.1609/aaai.v36i1.19999}
}

@inproceedings{Pavlakos,
  title={Learning to estimate {3D} human pose and shape from a single color image},
  author={Pavlakos, Georgios and Zhu, Luyang and Zhou, Xiaowei and Daniilidis, Kostas},
  booktitle=CVPR,
  pages={459-468},
  year={2018},
  doi={http://doi.org/10.1109/CVPR.2018.00055}
}

@inproceedings{NBF,
  title={{Neural Body Fitting}: Unifying deep learning and model based human pose and shape estimation},
  author={Omran, Mohamed and Lassner, Christoph and Pons-Moll, Gerard and Gehler, Peter and Schiele, Bernt},
  booktitle={3DV},
  pages={484-494},
  doi={http://doi.org/10.1109/3DV.2018.00062},      
  year={2018}
}

@inproceedings{pymaf,
  title={{PyMAF}: {3D} Human Pose and Shape Regression with Pyramidal Mesh Alignment Feedback Loop},
  author={Zhang, Hongwen and Tian, Yating and Zhou, Xinchi and Ouyang, Wanli and Liu, Yebin and Wang, Limin and Sun, Zhenan},
  booktitle=ICCV,
   doi={http://doi.org/10.1109/ICCV48922.2021.01125},
  pages={11426-11436},
  year={2021}
}

@inproceedings{xu2022vitpose,
  title={{ViTPose}: Simple vision transformer baselines for human pose estimation},
  author={Xu, Yufei and Zhang, Jing and Zhang, Qiming and Tao, Dacheng},
  booktitle=NIPS,
  year={2022}
}

@InProceedings{coco,
author="Lin, Tsung-Yi
and Maire, Michael
and Belongie, Serge
and Hays, James
and Perona, Pietro
and Ramanan, Deva
and Doll{\'a}r, Piotr
and Zitnick, C. Lawrence",
title="Microsoft COCO: Common Objects in Context",
booktitle=ECCV,
year="2014",
pages="740--755",
doi={https://doi.org/10.1007/978-3-319-10602-1_48}
}

@misc{sapiens,
    title={Sapiens: Foundation for Human Vision Models},
    author={Khirodkar, Rawal and Bagautdinov, Timur and Martinez, Julieta and Zhaoen, Su and James, Austin and Selednik, Peter and Anderson, Stuart and Saito, Shunsuke},
    year={2024},
    eprint={2408.12569},
    archivePrefix={arXiv},
    primaryClass={cs.CV},
    url={https://arxiv.org/abs/2408.12569}
}

@INPROCEEDINGS{Inter-part,
  author={Zhou, Yuxiao and Habermann, Marc and Habibie, Ikhsanul and Tewari, Ayush and Theobalt, Christian and Xu, Feng},
  booktitle=CVPR, 
  title={Monocular Real-time Full Body Capture with Inter-part Correlations}, 
  year={2021},
  volume={},
  number={},
  pages={4809-4820},

  doi={10.1109/CVPR46437.2021.00478}}

@INPROCEEDINGS{hdnet,
  author={Jafarian, Yasamin and Park, Hyun Soo},
  booktitle=CVPR, 
  title={Learning High Fidelity Depths of Dressed Humans by Watching Social Media Dance Videos}, 
  year={2021},
  volume={},
  number={},
  pages={12748-12757},
  doi={10.1109/CVPR46437.2021.01256}}

@INPROCEEDINGS{cloth4d,
  author={Zou, Xingxing and Han, Xintong and Wong, Waikeung},
  booktitle=CVPR, 
  title={CLOTH4D: A Dataset for Clothed Human Reconstruction}, 
  year={2023},
  volume={},
  number={},
  pages={12847-12857},
  doi={10.1109/CVPR52729.2023.01235}}

@article{dinov2,
  title={Dinov2: Learning robust visual features without supervision},
  author={Oquab, Maxime and Darcet, Timoth{\'e}e and Moutakanni, Th{\'e}o and Vo, Huy and Szafraniec, Marc and Khalidov, Vasil and Fernandez, Pierre and Haziza, Daniel and Massa, Francisco and El-Nouby, Alaaeldin and others},
  journal={arXiv preprint arXiv:2304.07193},
  year={2023}
}

@article{AIM,
  title={Scalable pre-training of large autoregressive image models},
  author={El-Nouby, Alaaeldin and Klein, Michal and Zhai, Shuangfei and Bautista, Miguel Angel and Toshev, Alexander and Shankar, Vaishaal and Susskind, Joshua M and Joulin, Armand},
  journal={arXiv preprint arXiv:2401.08541},
  year={2024}
}

@INPROCEEDINGS{Virtual_Markers,
  author={Ma, Xiaoxuan and Su, Jiajun and Wang, Chunyu and Zhu, Wentao and Wang, Yizhou},
  booktitle=CVPR, 
  title={3D Human Mesh Estimation from Virtual Markers}, 
  year={2023},
  volume={},
  number={},
  pages={534-543},
  
  doi={10.1109/CVPR52729.2023.00059}}

@INPROCEEDINGS{EMDB,
  author={Kaufmann, Manuel and Song, Jie and Guo, Chen and Shen, Kaiyue and Jiang, Tianjian and Tang, Chengcheng and Zárate, Juan José and Hilliges, Otmar},
  booktitle=ICCV, 
  title={EMDB: The Electromagnetic Database of Global 3D Human Pose and Shape in the Wild}, 
  year={2023},
  volume={},
  number={},
  pages={14586-14597},
  doi={10.1109/ICCV51070.2023.01345}}

@InProceedings{3DCrowdNet,  
author = {Choi, Hongsuk and Moon, Gyeongsik and Park, JoonKyu and Lee, Kyoung Mu},  
title = {Learning to Estimate Robust {3D} Human Mesh from In-the-Wild Crowded Scenes},  
booktitle = CVPR,
doi={http://doi.org/10.1109/CVPR52688.2022.00153},
year = {2022} ,
pages={1465-1474}

}

@inproceedings{3dpw,
author = {von Marcard, Timo and Henschel, Roberto and Black, Michael J. and Rosenhahn, Bodo and Pons-Moll, Gerard},
title = {Recovering Accurate {3D} Human Pose in the Wild Using IMUs and a Moving Camera},
year = {2018},
booktitle = ECCV,
pages = {614–631},
doi = {http://doi.org/10.1007/978-3-030-01249-6\_37},
numpages = {18}
}

@inproceedings{mpi-inf-3dhp,  
author={Mehta, Dushyant and Rhodin, Helge and Casas, Dan and Fua, Pascal and Sotnychenko, Oleksandr and Xu, Weipeng and Theobalt, Christian}, 
booktitle={3DV},  
title={Monocular {3D} Human Pose Estimation in the Wild Using Improved CNN Supervision},  
year={2017}, 
doi={http://doi.org/10.1109/3DV.2017.00064},
 volume={},  number={},  
pages={506-516},  
}

@article{survey,
  title={Recovering {3D} Human Mesh from Monocular Images: A Survey},
  author={Yating Tian and Hongwen Zhang and Yebin Liu and Limin Wang},
  journal=PAMI,
  year={2023},
  volume={45},
  number={12},
  pages={15406 - 15425},
  doi={10.1109/TPAMI.2023.3298850}
}

@article{pymaf-X,
title={{PyMAF-X}: Towards Well-Aligned Full-Body Model Regression From Monocular Images},
  author={Zhang, Hongwen and Tian, Yating and Zhang, Yuxiang and Li, Mengcheng and An, Liang and Sun, Zhenan and Liu, Yebin},
  journal=PAMI, 
    volume={},
  number={},
  pages={1-16},
  year={2023},
  doi={http://doi.org/10.1109/TPAMI.2023.3271691}}

@InProceedings{STRAPS,
               author = {Sengupta, Akash and Budvytis, Ignas and Cipolla, Roberto},
               title = {Synthetic Training for Accurate 3D Human Pose and Shape Estimation in the Wild},
               booktitle = BMVC,
               month = {September},
               year = {2020}                         
}

@article{ShapeBoost, 
    title={ShapeBoost: Boosting Human Shape Estimation with Part-Based Parameterization and Clothing-Preserving Augmentation}, 
    volume={38}, 
    url={https://ojs.aaai.org/index.php/AAAI/article/view/27841}, 
    DOI={10.1609/aaai.v38i2.27841},  
    number={2}, 
    journal=AAAI,
    author={Bian, Siyuan and Li, Jiefeng and Tang, Jiasheng and Lu, Cewu}, 
    year={2024}, 
    month={Mar.},
    pages={828-836} }

@InProceedings{shapy,
    author    = {Choutas, Vasileios and M\"uller, Lea and Huang, Chun-Hao P. and Tang, Siyu and Tzionas, Dimitrios and Black, Michael J.},
    title     = {Accurate 3D Body Shape Regression Using Metric and Semantic Attributes},
    booktitle = CVPR,
    month     = {June},
    year      = {2022},
    pages     = {2718-2728},
 doi={10.1109/CVPR52688.2022.00274}
}

@INPROCEEDINGS{tokenhmr,
  author={Dwivedi, Sai Kumar and Sun, Yu and Patel, Priyanka and Feng, Yao and Black, Michael J.},
  booktitle=CVPR, 
  title={TokenHMR: Advancing Human Mesh Recovery with a Tokenized Pose Representation}, 
  year={2024},
  volume={},
  number={},
  pages={1323-1333},
  doi={10.1109/CVPR52733.2024.00132}}

@InProceedings{isair,
author="Liu, Leyuan
and Gao, Yunqi
and Sun, Jianchi
and Chen, Jingying",
editor="Lu, Huimin
and Cai, Jintong",
title="Single-Image 3D Human Pose and Shape Estimation Enhanced by Clothed 3D Human Reconstruction",
booktitle="Artificial Intelligence and Robotics",
year="2024",
publisher="Springer Nature Singapore",
address="Singapore",
pages="33--44",
isbn="978-981-99-9109-9"
}

@article{ms,
  title={Single-image clothed 3D human reconstruction guided by a well-aligned parametric body model},
  author={Liu, Leyuan and Gao, Yunqi and Sun, Jianchi and Chen, Jingying},
  journal={Multimedia Systems},
  volume={29},
  number={3},
  pages={1579--1592},
  year={2023},
  publisher={Springer}
}

\clearpage
\appendix

\section{Network Details}
\begin{figure}[htp]
  \centering
  \includegraphics[width=0.45\textwidth]{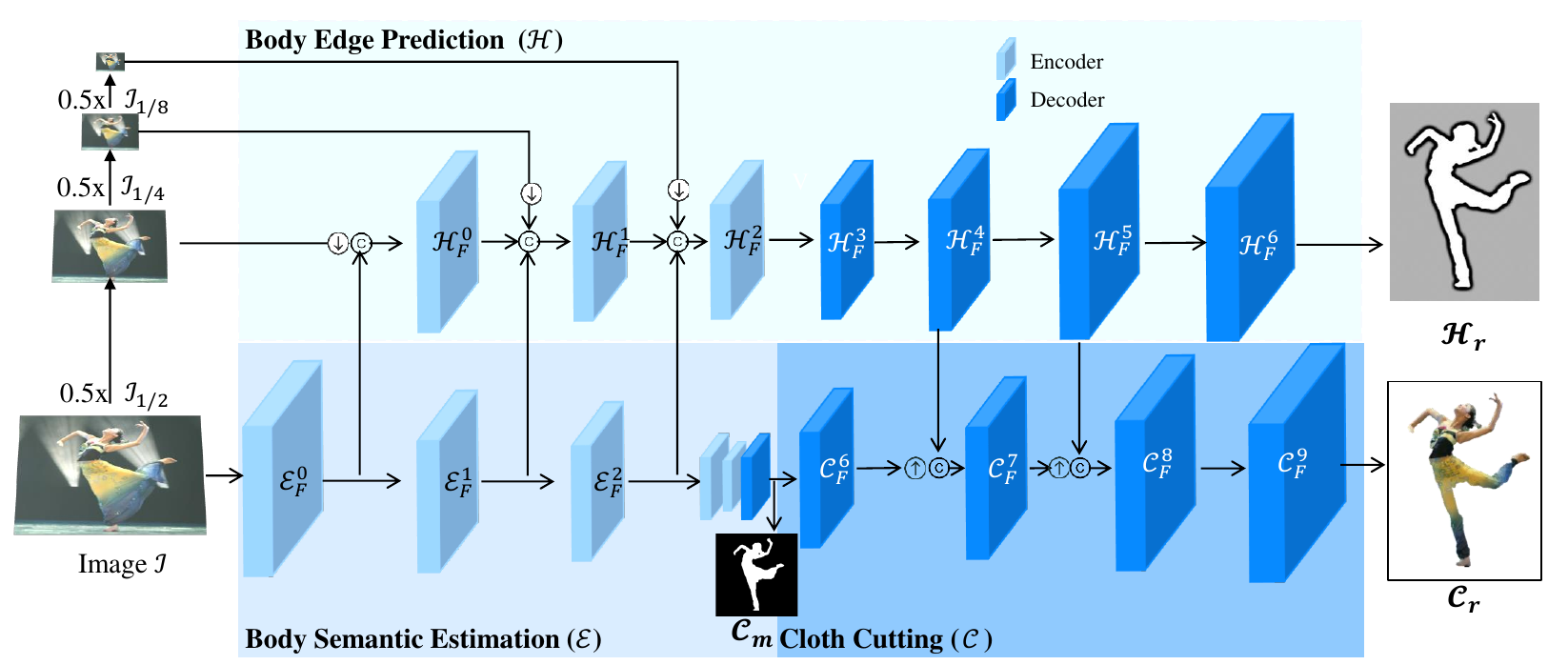}
  \caption{The clothing tailoring module's network details are shown. }
  \Description{}
  \label{sup_SEdetails}
\end{figure}

The clothing tailoring module consists of three parts: body semantic estimation ($\mathcal{E}$), edge prediction ($\mathcal{H}$), and Cloth Cutting ($\mathcal{C}$). Specifically, $\mathcal{E}$ includes five encoding layers and one decoding layer; $\mathcal{H}$ contains three encoding layers and four decoding layers; and $\mathcal{C}$ comprises four decoding layers. Each encoding and decoding layer is composed of convolutional layers and ReLU functions. The network details are shown in Fig.~\ref{sup_SEdetails}.

\section{Web Applications}

\begin{figure}[htp]
  \centering
  \includegraphics[width=0.38\textwidth]{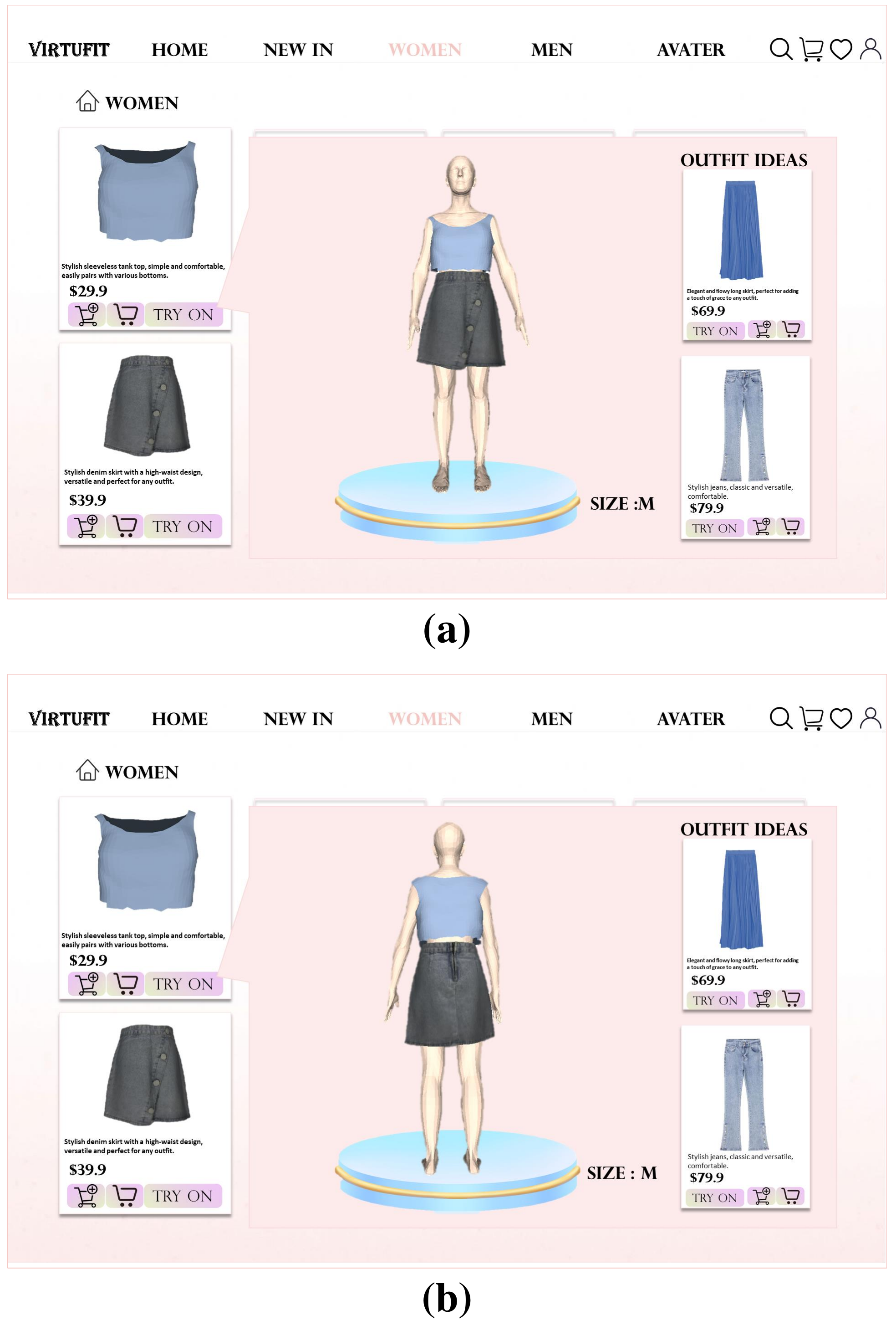}
  \caption{Web application based on ClothHMR. (a) and (b) are two webpages showing the 3D model rotated from different viewpoints. }
  \Description{}s
  \label{sup_web}
\end{figure}
\begin{figure}[htp]
  \centering
  \includegraphics[width=0.38\textwidth]{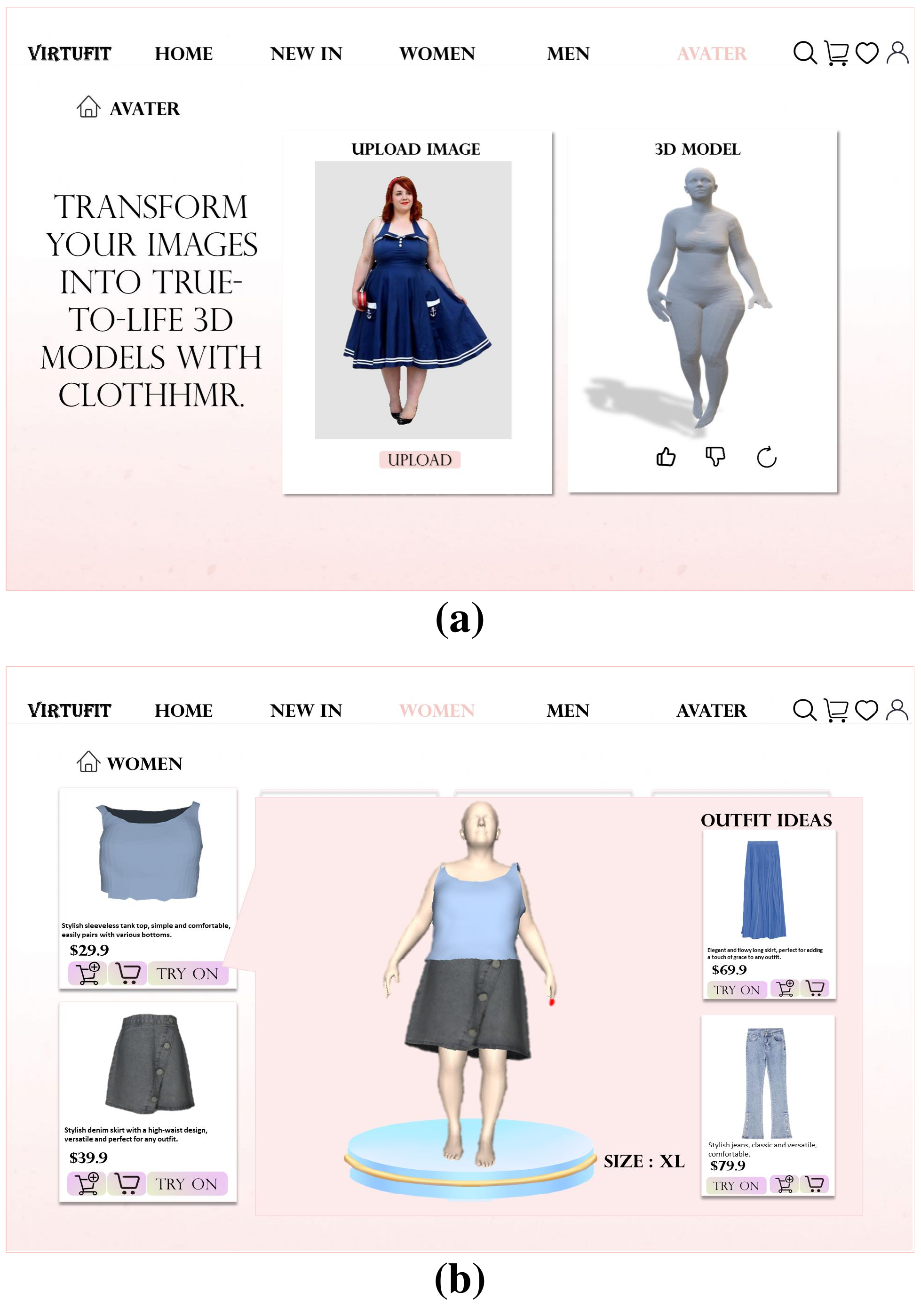}
  \caption{Web application based on ClothHMR. (a) presents another example with a different shape, and (b) displays the virtual try-on webpage. }
  \Description{}
  \label{sup_web2}
\end{figure}

Fig.~\ref{sup_web} displays multiple perspectives of the 3D model applied in the virtual fitting webpage using ClothHMR. Users can rotate the model in a full 360-degree view. Fig.~\ref{sup_web2} shows the body shape and pose estimation results of individuals with different body shapes, along with the clothing fitting outcomes. Additionally, ClothHMR can reconstruct the human model wearing any clothing. This approach significantly increases user engagement and improves the overall shopping experience.

\section{Qualitative Comparisons.}
\begin{figure*}
  \centering
  \includegraphics[width=0.9\textwidth]{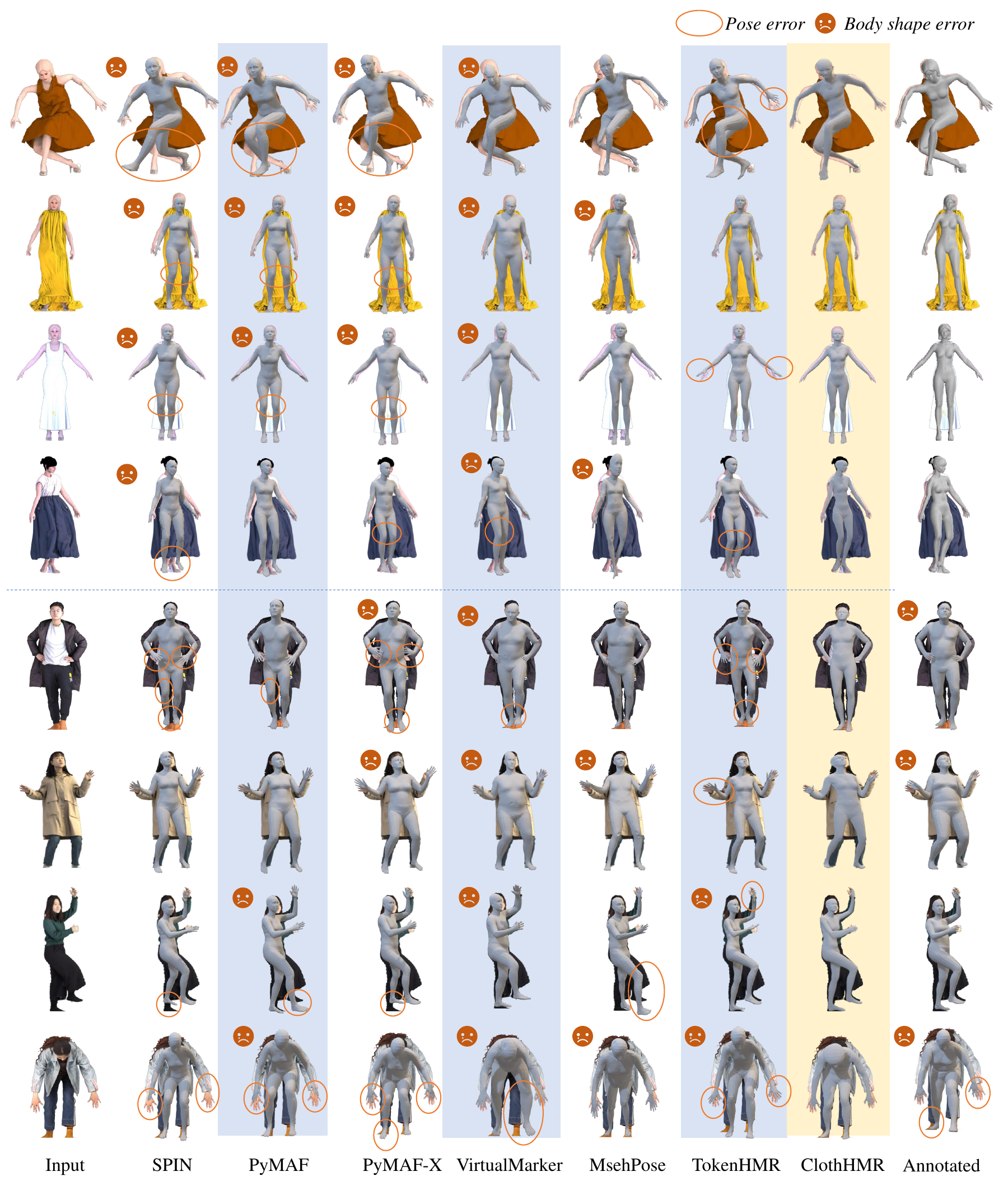}
  \caption{Visual comparison with other SOTA methods on Cloth4D and THuman2.0 datasets. ClothHMR can accurately estimate the pose and shape information of the body under loose clothing.}
  \Description{}
  \label{sup_cloth4d}
\end{figure*}

\begin{figure*}
  \centering
  \includegraphics[width=\textwidth]{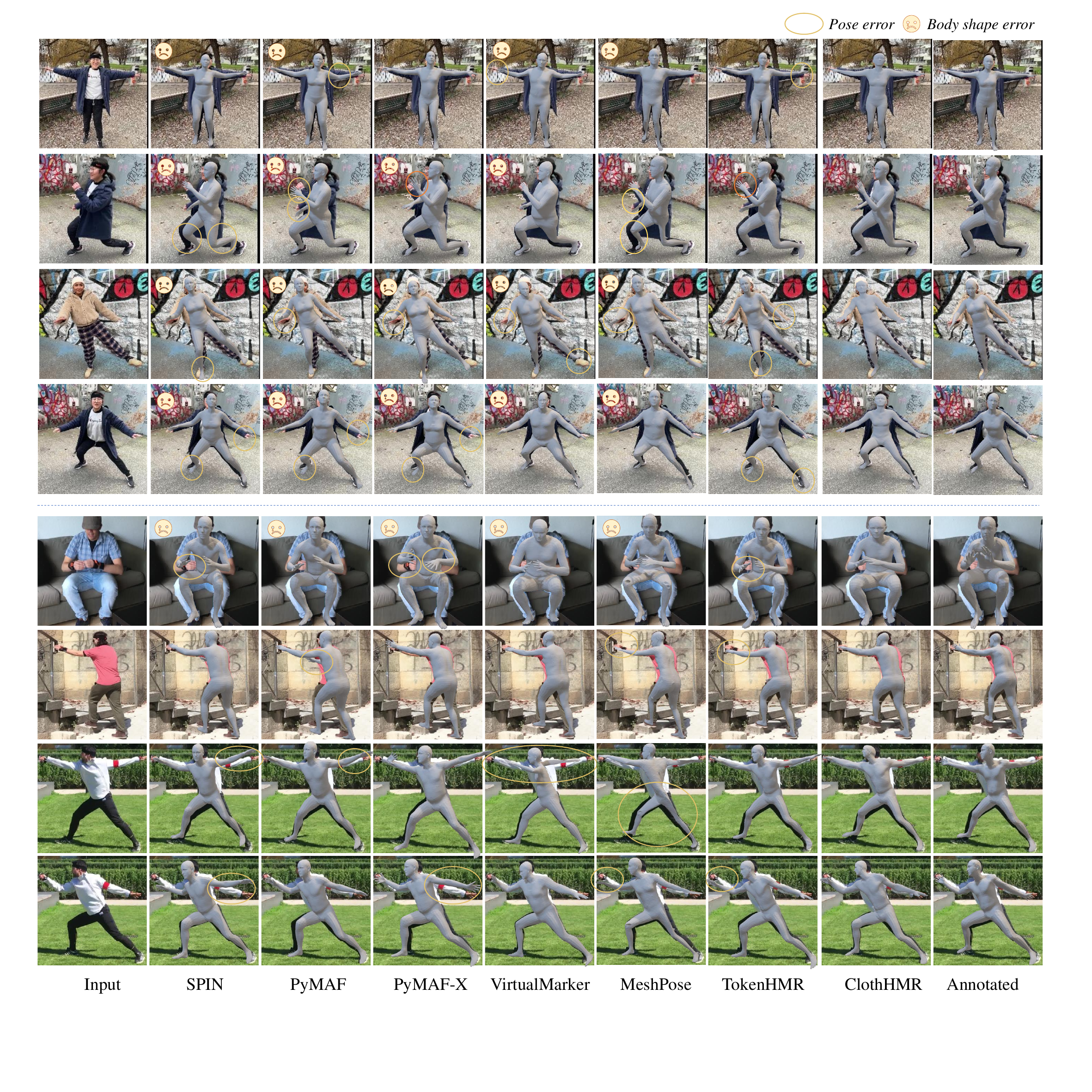}
  \caption{Visual comparison with other SOTA methods on EMDB and 3DPW datasets. ClothHMR can accurately estimate the pose and shape information of the body under loose coats and fitted clothing.}
  \Description{}
  \label{sup_emdb}
\end{figure*}

Fig.~\ref{sup_cloth4d} shows a qualitative comparison between our method and other advanced methods (SPIN~\cite{SPIN}, PyMAF~\cite{pymaf}, PyMAF-X~\cite{pymaf-X}, VirtualMarker~\cite{Virtual_Markers}, and TokenHMR~\cite{tokenhmr}) on the THuman2.0 and Cloth4D datasets. The results demonstrate that our method excels in pose accuracy and shape personalization. Meanwhile, Fig.~\ref{sup_emdb} compares our method with other advanced methods on the EMDB and 3DPW datasets, further proving the effectiveness of our method.

\begin{figure*}
  \centering
  \includegraphics[width=\textwidth]{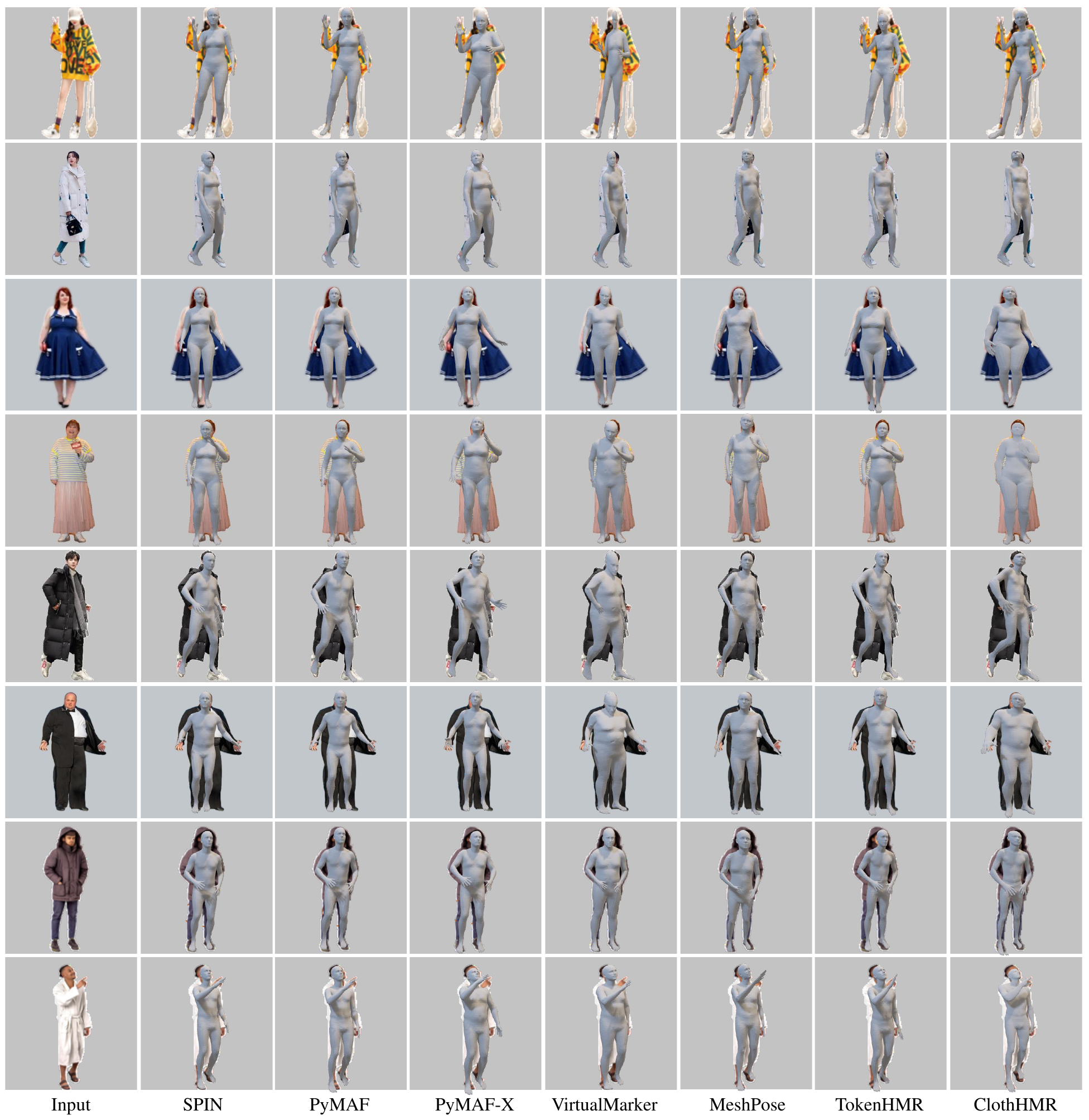}
  \caption{Visual comparison with other SOTA methods on in-the-wild loose clothing datasets. ClothHMR can accurately estimate the body shape information under loose coats and precisely fit the height and body type.
 }
  \label{sup_losses_cloth}
\end{figure*}

\begin{figure*}
  \centering
  \includegraphics[width=\textwidth]{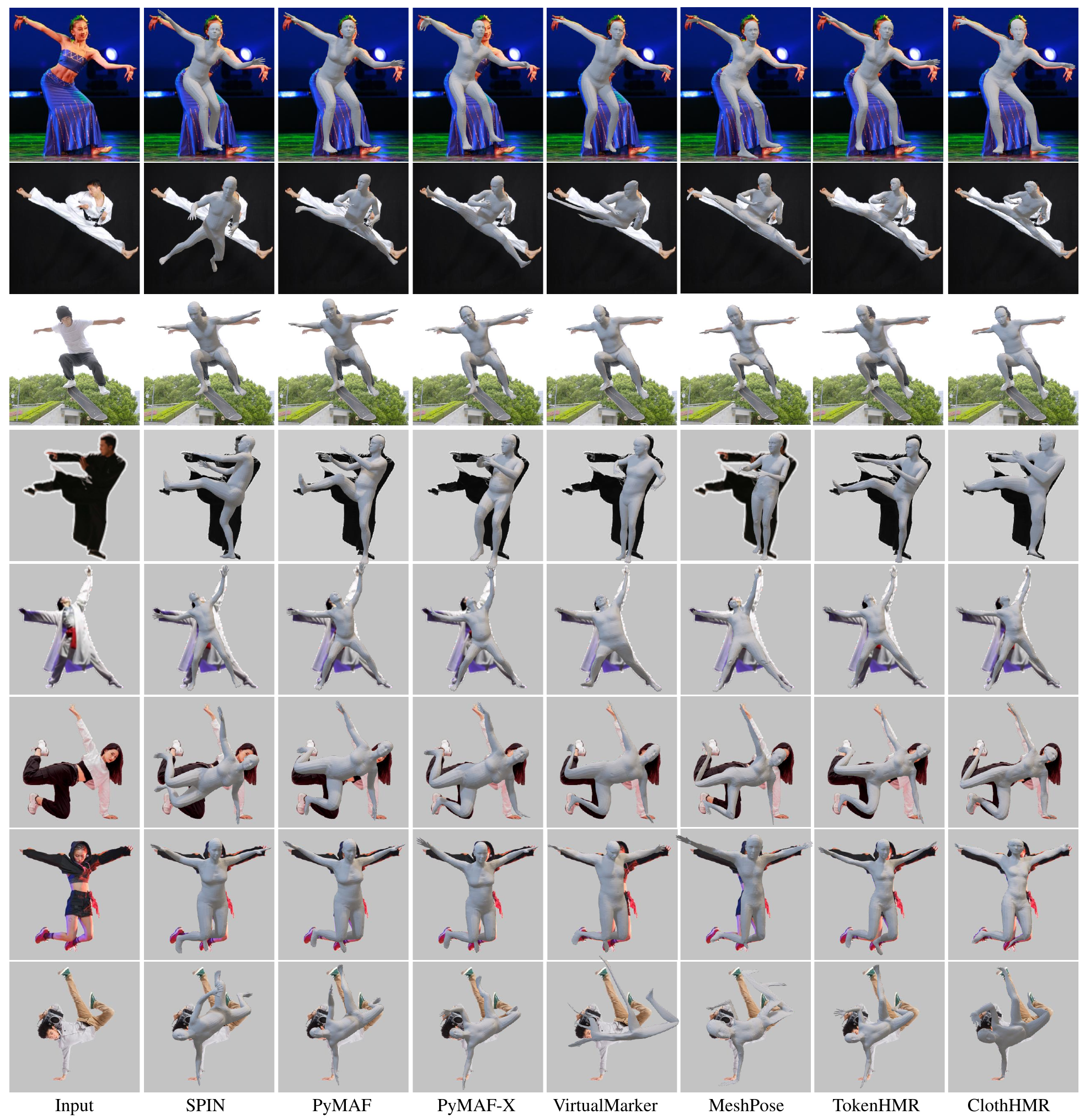}
  \caption{Visual comparison with other SOTA methods on in-the-wild challenge poses datasets. ClothHMR can accurately fit the complex pose of the body under loose coats.}
  \Description{}
  \label{sup_challenge_pose}
\end{figure*}

Additionally, Fig.~\ref{sup_losses_cloth} and Fig.~\ref{sup_challenge_pose} respectively compare our method with other advanced methods on in-the-wild loose clothing datasets and complex pose clothing datasets. ClothHMR performs exceptionally well on these challenging in-the-wild datasets, accurately estimating body poses and shapes under clothing, demonstrating strong generalization capabilities. The experimental results show that our method performs excellently on standard datasets and exhibits robust performance in practical applications.



\end{document}